\documentclass[lettersize,journal]{IEEEtran}
\usepackage{amsmath,amsfonts}
\usepackage{algorithmic}
\usepackage{algorithm}
\usepackage{array}
\usepackage[caption=false,font=normalsize,labelfont=sf,textfont=sf]{subfig}
\usepackage[colorlinks=true, linkcolor=cyan, citecolor=cyan, urlcolor=cyan]{hyperref}
\usepackage{textcomp}
\usepackage{stfloats}
\usepackage{url}
\usepackage{verbatim}
\usepackage{graphicx}
\usepackage{cite}
\usepackage{cleveref}
\usepackage{amssymb} 
\usepackage{multirow}
\usepackage{makecell}
\usepackage{booktabs}
\usepackage{pifont}
\begin{document}

\title{SM3D: Mitigating Spectral Bias and Semantic Dilution in Point Cloud State Space Models}

\author{
	Bin Liu, 
	Chunyang Wang, 
	Xuelian Liu,  
	Bo Xiao, 


%
	
}



\maketitle

\begin{abstract}
Point clouds are a fundamental 3D data representation that underpins various computer vision tasks.
Recently, Mamba has demonstrated strong potential for 3D point cloud understanding. 
However, existing approaches primarily focus on point serialization, overlooking a more fundamental limitation: State Space Models (SSMs) inherently exhibit a spectral low-pass bias arising from their recursive formulation.  
In serialized point clouds, this bias is particularly detrimental, as it suppresses high-frequency geometric structures and dilutes semantic discriminability across deep layers. 
To address these limitations, we propose SM3D, a spectral-aware framework designed to synergistically preserve geometric fidelity and semantic consistency.  
First, a Geometric Spectral Compensator (GSC) is introduced to counteract the low-pass bias by explicitly injecting graph-guided high-frequency components via local Laplacian analysis, thereby restoring structural sensitivity.
Second, we design a Semantic Coherence Refiner (SCR) to rectify semantic drift through frequency-aware channel recalibration. 
To balance theoretical precision and computational efficiency, SCR is instantiated via two pathways: an exact Laplacian eigendecomposition (SCR-L) and a linear-complexity Chebyshev polynomial approximation (SCR-C).
Extensive experiments demonstrate that SM3D achieves state-of-the-art performance, including 96.0\% accuracy on ModelNet40 and 86.5\% mIoU on ShapeNetPart, validating its effectiveness in mitigating spectral low-pass bias and semantic dilution.
Our code is available at \url{https://github.com/L1277471578/SM3D} 
\end{abstract}


\section{Introduction}

\IEEEPARstart{3D} point clouds constitute a fundamental data representation for numerous applications including autonomous driving \cite{DAM}, robotics \cite{CliReg}, and VR/AR systems \cite{Weber2024}. While their fine-grained geometric structure enables accurate perception, their unordered and irregular nature poses significant challenges for efficient sequential modeling. Recently, Mamba \cite{Mamba} has shown promise as an alternative to Transformer \cite{Transformer} architectures for 3D perception, as it offers linear computational complexity and strong long-sequence modeling capabilities. However, adapting Mamba to 3D point clouds typically requires flattening 3D topology into 1D token sequences. Consequently, existing Mamba-based approaches predominantly focus on optimizing serialization strategies to preserve spatial locality within the sequence \cite{pointramba}, \cite{PCM}, \cite{GridMamba}, \cite{Explor-lu2025}.

Despite these efforts, we argue that focusing solely on serialization overlooks a more fundamental and rarely discussed limitation: the intrinsic spectral low-pass bias of State Space Models (SSMs). Owing to their recursive formulation, SSMs inherently act as low-pass filters, systematically attenuating high-frequency components during state transitions \cite{Mamba}. This bias is particularly detrimental in serialized point clouds. Serialization inevitably disperses local geometric topology \cite{Explor-lu2025}, \cite{ptv3}, \cite{strumamba3d}, causing fine-grained structural details to manifest as high-frequency spectral fluctuations in the 1D sequence. The SSM's low-pass nature inadvertently suppresses these critical cues, leading to structural blurring. Importantly, this phenomenon originates from the spectral properties of the SSMs themselves, meaning it cannot be fully resolved by improved scanning strategies alone.

\begin{figure}[!t]
	\centering
	\includegraphics[width=0.495\textwidth]{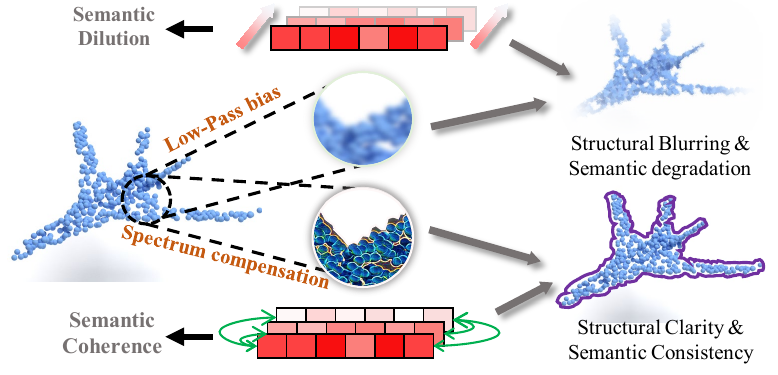}
	\caption{ Illustration of SM3D mitigating spectral low-pass bias and semantic dilution. The upper branch illustrates feature blurring caused by the spectral bias of SSMs and semantic dilution. The lower branch highlights the high frequency spectral compensation and semantic coherence.
	}
	\label{fig1}
\end{figure}

This spectral bias further cascades into semantic dilution during deep propagation. As high-frequency geometric components are suppressed, feature representations progressively lose discriminative power. Consequently, spectral bias induces continuous feature drift \cite{lee2024}, where accumulated smoothing under channel-isolated processing prevents the model from maintaining semantic coherence. This results in progressive divergence and blurring of semantic representations in deeper layers, ultimately undermining the overall representation.

To address these challenges, we propose SM3D, a spectral-aware framework designed to mitigate spectral bias and semantic dilution in point cloud understanding. To our knowledge, this is the first work to analyze point cloud Mamba models for spectral low-pass bias. Specifically, we introduce a Geometric Spectral Compensator (GSC) that explicitly performs local Laplacian-based analysis to extract and inject graph-guided high-frequency residuals. This mechanism effectively counterbalances the low-pass tendency of SSMs, restoring sensitivity to fine-grained structures. Furthermore, we design a Semantic Coherence Refiner (SCR), a spectral–spatial interaction component that performs frequency-aware channel recalibration to maintain global semantic consistency. To balance accuracy and efficiency, SCR is realized via two complementary pathways: an exact spectral formulation based on Laplacian eigendecomposition (SCR-L), and a Chebyshev polynomial–based spectral approximation (SCR-C) that achieves linear computational complexity. Together, these components ensure that SpecMamba3D preserves both geometric fidelity and semantic integrity throughout the deep SSM propagation process, as illustrated in Fig.\ref{fig1}.

The main contributions of this work are summarized as follows:
\begin{itemize}

	\item	We identify and analyze the intrinsic spectral low-pass bias of SSMs when applied to serialized point clouds, and reveal its role in suppressing high-frequency geometric information and causing semantic dilution.
	
	\item	We propose a Geometric Spectral Compensator (GSC) that explicitly injects graph-guided high frequency responses via local Laplacian analysis, effectively mitigating spectral bias in Mamba-based point cloud models without introducing excessive computational overhead.
	
	\item	We design a Semantic Coherence Refiner (SCR), which maintains semantic consistency across frequency bands during deep propagation. It admits efficient instantiations under different spectral formulations, enabling a flexible trade-off between precision and efficiency.
	
	\item	Extensive experiments on ModelNet40, ScanObjectNN, ShapeNetPart, and S3DIS benchmarks demonstrate that SM3D achieves state-of-the-art performance with a favorable accuracy–efficiency trade-off, confirming the importance of addressing spectral bias and semantic dilution in point cloud SSMs.

\end{itemize}

\section{Related Work}
\label{Related Work}

\subsection{Spectral Methods in Point Cloud Deep Learning}

Effective point cloud understanding relies on characterizing local geometric structures, such as shape variations and neighborhood relationships. Representative methods explicitly model these through hierarchical grouping, dynamic graphs, and spatial weighting \cite{KPConv}, \cite{PointNet2}, \cite{dgcnn}, \cite{DPC}, \cite{SpiderCNN}. While effective, these spatial methods often struggle to extract intrinsic geometric properties efficiently.

Spectral-domain analysis provides a complementary perspective by modeling point clouds in the graph framework. PointWavelet \cite{PointWavelet} exploits graph wavelet transforms to capture multi-scale spectral features, while S$^2$ANet \cite{S2ANet} integrates spectral responses with spatial local aggregation for enhanced geometric modeling. More recent studies, including Spectral-GANs \cite{Spectral-GANs} and PointGST \cite{pointgst}, further demonstrate the effectiveness of spectral constraints in high-resolution shape generation and parameter-efficient fine-tuning, highlighting the growing potential of spectral representations for point cloud understanding.


\subsection{Transformer and Mamba for Point Clouds}

Transformers have achieved state-of-the-art performance in 3D perception by utilizing self-attention to model long-range dependencies \cite{Pointgpt}, \cite{point-femae}, \cite{ptv3}, \cite{point-mae}. However, the quadratic complexity of global attention restricts scalability. While sparse and patch-based variants alleviate computational costs \cite{centerformer}, \cite{PatchFormer}, \cite{DSVT}, \cite{flatformer} they often compromise the integrity of geometric structures.

Recent studies have analyzed the spectral bias of SSMs in 1D signal processing \cite{Wave-U-Mamba} and 2D vision \cite{XYScanNet}. However, their implications for irregular 3D point clouds remain unexplored. Existing Mamba-based point cloud methods, including PointMamba \cite{PointMamba}, PCM \cite{PCM}, PoinTramba \cite{pointramba}, and DM3D \cite{DM3D}. primarily focus on optimizing point serialization strategies to improve spatial continuity. While recent works such as Mamba3D \cite{Mamba3D} and StruMamba3D \cite{strumamba3d} incorporate local structural features to enhance geometric details, they do not address the core spectral bias within the SSMs, resulting in progressive attenuation of high-frequency geometric information across deep layers.

In summary, existing Mamba-based point cloud methods emphasize improved serialization and local aggregation, whereas SM3D fundamentally differs in motivation and design by explicitly compensating spectral bias and rectifying semantic dilution.


\section{Method}
\label{Method}

\subsection{Problem Formulation}
\label{problem}

Mamba  has demonstrated strong capability via its input-dependent selection mechanism. The core Selective State Space Model (S6) \cite{Mamba} operates through discrete recurrence:
\begin{align}
	\label{SSM}
	{{h}_{t}}={{\bar{A}}_{t}}{{h}_{t-1}}+{{\bar{B}}_{t}}{{x}_{t}},{{y}_{t}}={{\bar{C}}_{t}}{{h}_{t}}
\end{align}
where parameters $\bar{A}_t$, $\bar{B}_t$, ${\bar{C}}_t$ are dynamically generated from the input $x_t$.

\textit{Low-Pass Bias in Mamba. }
The input-dependent nature of Mamba renders it a time-varying system, precluding classical frequency response analysis via Z-transforms, as transfer functions are well-defined only for time-invariant systems. Consequently, we examine its spectral properties via the effective impulse response in the time domain, which is more suitable for point clouds from a graph signal processing perspective.

The recurrence in Eq.(\ref{SSM}) allows the state evolution to be expressed as a cumulative aggregation of historical input tokens. Assuming an initial state $h_0=0$, the output at time $t$ can be expanded as:
\begin{align}
{{y}_{t}}={{\bar{C}}_{t}}\sum\limits_{k=1}^{t}{\left( \prod\limits_{m=k+1}^{t}{{{{\bar{A}}}_{m}}} \right){{{\bar{B}}}_{k}}}{{x}_{k}}={{\bar{C}}_{t}}\sum\limits_{k=1}^{t}{\phi (t,k)}{{x}_{k}}
\end{align}
where $\phi(t,k)$  represents the effective impulse response of input token $x_k$  at time $t$. 

Crucially, for system stability, the dynamic discretization state matrix ${{\bar{A}}_{t}}$ must satisfy the condition that its spectral radius $\rho ({{\bar{A}}_{t}})<1, \forall t$ \cite{Gu2022}. Therefore, there exists a constant $\varphi \in (0, 1)$ such that $\lVert{{\bar{A}}_{t}}\rVert \le \varphi, \forall t$. This implies $\lVert \phi(t,k) \rVert \le {{\varphi }^{t-k}}\lVert{{\bar{B}}_{k}}\rVert$. The magnitude of $\phi(t,k)$ decays exponentially with the temporal distance $t-k$. This indicates that Mamba behaves as a low-pass–biased operator, preserving low-frequency trends while suppressing high-frequency variations.

A point cloud $\mathcal{P}$ can be viewed as a spatial graph in which high-frequency spectral components encode critical geometric details (e.g., edges and sharp curvature). Serializing $\mathcal{P}$ separates neighboring points, thereby dispersing high-frequency components. Due to the low-pass bias, such components are systematically attenuated during sequential propagation, leading to structural blurring.

\begin{figure*}[!ht]
	
	\centering
	\includegraphics[width=0.9\textwidth]{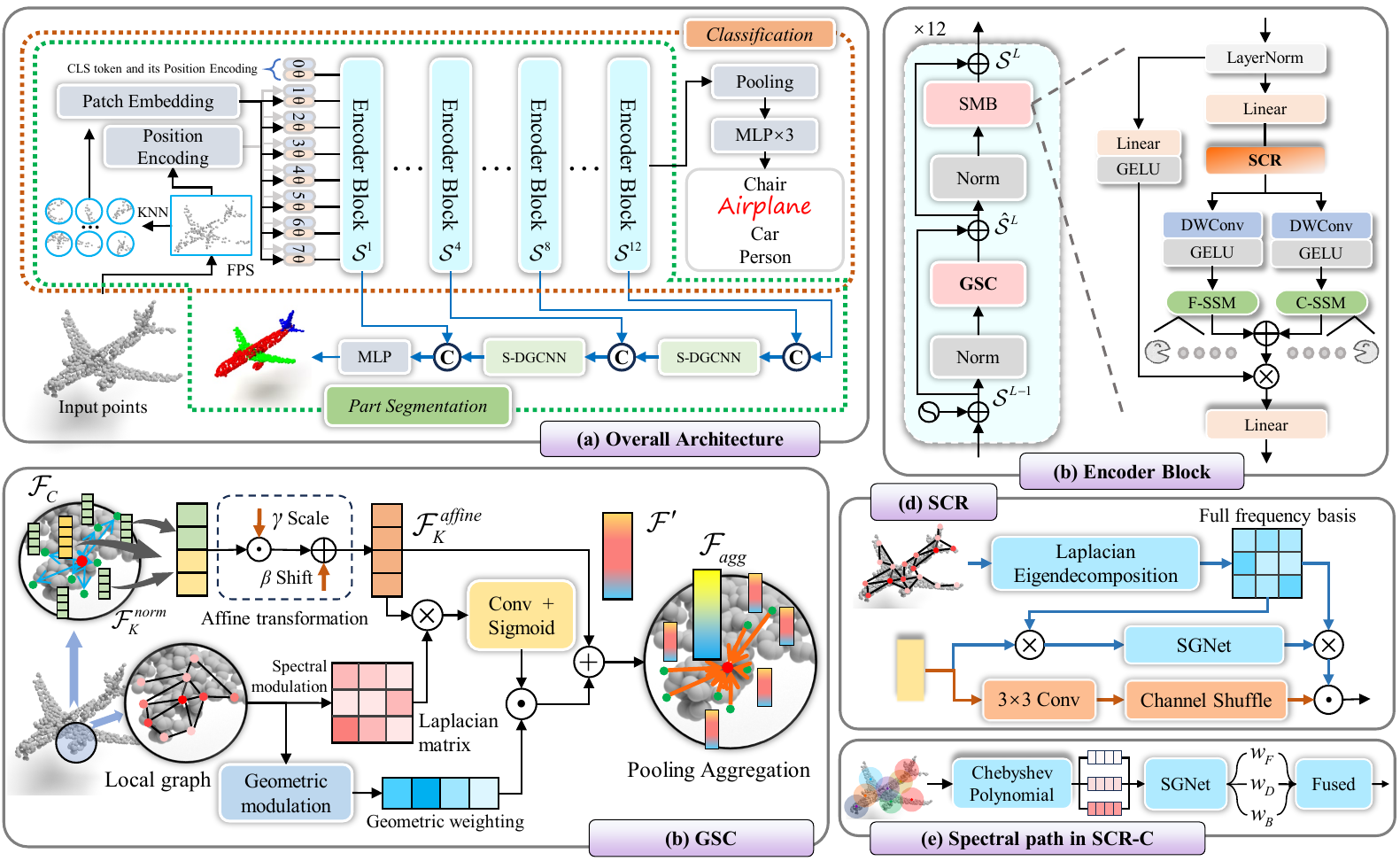}
	\caption{Overview of SM3D.
		(a) Overall architecture showing the embedding,encoder,and decoder structures. 
		(b) Encoder Block core consists of GSC, SCR and two SSM branches: standard forward SSM \cite{PointMamba} (F‑SSM) and channel‑flipped backward SSM  \cite{Mamba3D} (C-SSM). 
		(c) GSC, extracts high frequency geometric residual.
		(d) SCR, constructs global spectral anchors. 
		(e) The spectral path in SCR‑C, approximates spectral filtering via Chebyshev polynomials.
		Symbols: © denotes channel concatenation, $\oplus$ residual addition, $\otimes$ matrix multiplication, and $\odot$ element-wise multiplication. }
	\label{architecture}
\end{figure*} 

\textit{Deep Semantic Dilution.}
The low-pass bias not only affects local details but also induces a degradation of semantic consistency during deep propagation. Considering a stacked SSM with $L$ layers, the contribution of an input $x_{i}^{(0)}$ to the final hidden state $h_{t}^{(L)}$  can be expressed as a cumulative contribution of transition matrices:
\begin{align}
	h_t^{(L)} =
\sum_{k=1}^{t}
\left(
\prod_{l=1}^{L}
\prod_{m=k+1}^{t}
\bar{A}_{m}^{(l)}
\right)
\bar{B}_{k}^{(l)} x_k^{(0)}
\end{align}
where $\bar{A}_{m}^{(l)}$ and $\bar{B}_{k}^{(l)}$ denote the transition and input matrices at layer $l$ and token index $m$ or $k$, respectively.

This expression reveals that the influence of point $k$ on point $t$ is constrained by the sequential distance $\|t-k\|$ and network depth $L$, causing semantic associations captured in shallow layers to progressively dilute as network depth increases.Distinct from the over-smoothing induced by spectral low-pass bias, this degradation accumulates across layers as feature drift. Moreover, SSMs process feature channels independently, which lacks a mechanism to cross-validate semantic information across channels, which further exacerbating global semantic coherence degradation.

\vspace{-2mm}

\subsection{Overview}
\label{overview}
To address the challenges in Section \ref{problem}, we propose SM3D, which compensates for geometry-aware high-frequency components via the GSC and enforces semantic recalibration using the SCR. 

\textit{Patch Embedding.}
As shown in Fig.\ref{architecture}a, given an input point cloud $\mathcal{P}$, we first transform the unordered data into a discrete token sequence compatible with SSMs. Specifically, we sample $N$ centroids $\mathcal{P}_C \in \mathbb{R}^{N \times 3}$ via Farthest Point Sampling (FPS) and $K$ group local neighborhoods using K-Nearest Neighbors (KNN) to form patches. A lightweight PointNet projects these patches to generate token embedding $\mathcal{F}_C \in \mathbb{R}^{N \times D}$. A learnable class token $x_{cls}$ and patch positional encodings are added, forming initialized sequence $\mathcal{S}^0 \in \mathbb{R}^{(N+1) \times D}$, where $D$ is the embedding dimension.

\textit{Encoder Block.}
SM3D’s backbone is built upon 12 stacked encoder blocks, each adopting a dual-stage residual architecture. As shown in Fig.\ref{architecture}(b), each block first employs the GSC to injecting graph-guided high-frequency components. The enhanced features are then processed by a Spectral Mamba Block (SMB) integrated with SCR, which calibrating channel semantics. Formally, the $L$-th encoder block is defined as:
\begin{align}
	& {{{\hat{\mathcal{S}}}}^{L}}=GSC\left( Norm({{\mathcal{S}}^{L-1}}) \right)+{{\mathcal{S}}^{L-1}}, \\ 
	& {{\mathcal{S}}^{L}}=SMB\left( Norm({{\widehat{\mathcal{S}}}^{L}}) \right)+{{\widehat{\mathcal{S}}}^{L}}  
\end{align}
where  $Norm(\cdot )$ and ${\mathcal{S}}^{L-1}$ denotes Layer Normalization and ($L-1$)-layer output, respectively.

\textit{Task Decoders.}
After the encode blocks, task heads are employed to decode the learned representations.  For classification, the final class token $x_{cls}$ is concatenated with the global max- and average-pooled features of $\mathcal{S}^{12}$ and projected via an MLP to get classification prediction results. For segmentation, we adopt a PointBERT-style decoder \cite{Point-BERT} with hierarchical feature propagation strategy. Multi-scale features from 1th, 4th, 8th, 12th layers are interpolated and refined using simplified DGCNN \cite{dgcnn} (S-DGCNN) to generate fine-grained segmentation predictions.

\vspace{-2mm}

\subsection{Geometric Spectral Compensator (GSC)}

GSC selectively injects high frequency spectral residuals into the token representations (see Fig.\ref{architecture}c).

\subsubsection{Local Graph Construction and Standardization}

Given the centroid set $\mathcal{P}_C$ and  features $\mathcal{F}_C$ defined in \cref{overview}, we first construct local geometric primitives. For each centroid ${{p}_{c}}\in {{\mathcal{P}}_{C}}$ , we retrieve its KNN to form a neighborhood ${{\mathcal{P}}_{g}}=\{{{p}_{c}}\}\cup {{\mathcal{N}}}({{p}_{c}})$, where  ${{\mathcal{N}}}({{p}_{c}})$ denotes the $K_G$  nearest neighbors of $p_c$  in Euclidean space.  The feature of point centroid $p_i$ pair is $\mathcal{F}_K \in \mathbb{R}^{N \times K_G \times D}$.

To ensure robustness against rigid transformations and scale variations, relative coordinates and feature differences are normalized by their standard deviation, goting $\mathcal{P}_{K}^{norm}$ and $\mathcal{F}_{K}^{norm}$ 
The normalization emphasizes local variations rather than absolute magnitudes, which is essential for consistent spectral analysis across patches. Then, we couple spatial and semantic information via a learnable affine transformation to generate the extended feature space $\mathcal{F}_{K}^{aff} \in \mathbb{R}^{N \times K_G \times 2D}$:
\begin{align}
	\mathcal{F}_{K}^{aff}=Concat(\mathcal{F}_{K}^{norm},{{\mathcal{F}}_{C}})\odot \gamma +\beta 
\end{align}
where $\gamma$, $\beta \in {{\mathbb{R}}^{2D}}$  are learnable scale and shift parameters. This mechanism local relations to initially couple spatial structure with features.

\subsubsection{High Frequency Extraction via Graph Laplacian}
To explicitly capture the high-frequency components, we construct an undirected weighted graph ${{\mathcal{G}}_{l}}=({{\mathcal{V}}_{l}},{{\mathcal{E}}_{l}},{\mathcal{W}_{l}})$ based on ${{\mathcal{P}}_{C}}$, where vertex set ${{\mathcal{V}}_{l}}=\left\{ {{v}_{l,k}} \right\}_{k=1}^{{{K}_{S}}},{{v}_{l,k}}\in {{\mathbb{R}}^{3}}$ consists of the $K_G$  nearest neighbors of ${{\mathcal{P}}_{C}}$.  Edges are established between all pairs of nodes within the local patch. To quantify the geometric topology of  ${{\mathcal{G}}_{l}}$, we compute a Gaussian adjacency matrix   ${\mathcal{W}_l}\in {{\mathbb{R}}^{{{K}_{G}}\times {{K}_{G}}}}$: 
\begin{align}
	& {{w}_{ij}^l}=\exp \left( -\frac{{{d}_{ij}}^{2}}{2{{\sigma }_{g}}^{2}} \right),\forall i\ne j \\ 
	& {{\sigma }_{l}}=\frac{1}{{{K}_{G}}^{2}}\sum\nolimits_{i=1}^{{{K}_{G}}}{\sum\nolimits_{j}^{{{K}_{G}}}{{{d}_{ij}}}} 
\end{align}
where the entry $w_{ij}^l$ in $\mathcal{W}_l$ indicates the weight of the edge between vertices $i$ and $j$, reflecting their similarity. ${{d}_{ij}}=||{{v}_{l,i}}-{{v}_{l,j}}|{{|}_{2}}$  denote Euclidean distance between vertices.  $\sigma_l$ is the local adaptive scaling parameters.

The normalized local graph Laplacian matrix \cite{Laplacian} $\mathcal{L}_l\in {{\mathbb{R}}^{{{K}_{G}}\times {{K}_{G}}}}$ is then given by:
\begin{align}
	\mathcal{L}_l=I-{{D}_{l}}^{-\frac{1}{2}}{\mathcal{W}_{l}}{{D}_{l}}^{-\frac{1}{2}}
	\label{L}
\end{align}
where $D_l$ is the degree matrix with diagonal element $\sum_j w_{ij}^l$.

In spectral graph theory, the operation $\mathcal{L}x$ acts as a high-pass filter, quantifying the discrepancy between a node and its neighbors \cite{Ortrga2018}. We apply this operator to the extended features $\mathcal{F}_{K}^{aff}$ to extract the high-frequency response $\mathcal{F}_{hf} = \mathcal{L}_l \mathcal{F}_{K}^{aff}$. This response explicitly encodes sharp geometric transitions that correspond to large eigenvalues of the Laplacian spectrum. Notedly, This operation that avoids complex spectral decomposition, still emphasizes high-frequency geometric variations, and preserving its filtering behavior. 

Not all high frequency signals are beneficial, we introduce learnable frequency selection gate to  select informative spectral bands:
\vspace{-2mm}
\begin{align}
	{\mathcal{F}'_{hf}}={{\sigma }}\left( Con{{v}_{1\times 1}}({\mathcal{F}_{hf}}) \right)
\end{align}
where  ${\sigma }$ is the Sigmoid activation function.

$\mathcal{F}'_{hf}$ represents the refined high-frequency residuals critical for structural discrimination, which correspond to feature components that are most susceptible to suppression under the intrinsic low-pass bias.

\subsubsection{Geometric Modulation and Pool Aggregation}
A key challenge in point cloud serialized modeling is that topological proximity is often disrupted. To ensure respects the original 3D structure, we employ geometric to modulation spectral restoration. 

Specifically, high-frequency cues preferentially derived from spatially proximate neighbors. Using the normalized distances, we compute geometry-aware modulation weights ${{w}_{geo}}\in {{\mathbb{R}}^{N\times K_G \times D}}$:
\begin{align}
	& {{w}_{geo}}=\exp \left( -\frac{|\mathcal{P}_{K}^{norm}{{|}_{2}}}{{{\sigma }_{geo}}+\varepsilon } \right) 
\end{align}
where ${{\sigma }_{geo}}$ is the local geometric weight scaling parameters.


These weights suppress contributions from distant points. The modulated high frequency residuals are then injected back into the feature stream via a residual connection:
\begin{align}
	{\mathcal{F}'}=\mathcal{F}_K^{aff}+\alpha \cdot ({\mathcal{F}'_{hf}}\odot {{W}_{geo}})
\end{align}
where $\mathcal{F}'$  is the final spectral-compensated feature. $\alpha$ is a learnable injection intensity parameter. ${{W}_{geo}}\in {{\mathbb{R}}^{N\times K_G \times 2D}}$ expands the weight $w_{geo}$ along the channel dimension.

This injection mechanism embeds spatial structure into feature channels, re-coupling geometric structure with the serialized token sequence.
Finally, we aggregate the compensated patch features into token by a Pooling strategy, and  projects the token back to dimension $D$ by a MLP, completing the compensation process.
%

\vspace{-2mm}

\subsection{Semantic Coherence Refiner (SCR)}


SCR integrating local spatial context with global spectral anchors to enforce cross-channel semantic integrity before recursive propagation. The architecture is detailed in Fig.\ref{architecture}d.

\subsubsection{Construction of Global Spectral Semantic Anchors}
The core objective of SCR is to extract permutation-invariant global semantics reference to rectify feature drift in deep propagation. 
We formulate this as a process that spectral filter $g(\mathcal{L})$ applies to input features $X \in \mathbb{R}^{N \times D}$, the spectrally calibrated representation $g(\mathcal{L})X$ serves as a semantic reference to recalibrate channel activations. We provide two distinct instantiations tailored for different trade-offs between precision and scalability.

\textit{SCR-L: Exact Spectral Anchoring via Laplacian Eigenbasis.}
For precision-critical scenarios with moderate point counts, we employ the exact Graph Fourier Transform (GFT). As shown in Fig.\ref{architecture}d, let $\mathcal{G}_g= (\mathcal{V}_g, \mathcal{E}_g, \mathcal{W}_g)$ denote the KNN graph constructed on the centroid set $\mathcal{P}_C$, where each node corresponds to a patch token. Based on pairwise distances $d_{ij}=\|p_i-p_j\|_2$, define the adjacency matrix $\mathcal{W}_g$ with entries $\mathcal{W}_{ij} = (d_{ij}/\eta_i + \delta_{ij})^{-1}$ \cite{pointgst}, where $\eta_i = \min_{k \ne i} d_{ik}$ is the nearest-neighbor distance and $\delta_{ij}$ is the Kronecker delta. This yields the Laplacian matrix $\mathcal{L}_g=I-D_{g}^{{}^{-1}\!\!\diagup\!\!{}_{2}\;}{\mathcal{W}_{g}}D_{g}^{{}^{-1}\!\!\diagup\!\!{}_{2}\;}$, with ${D}_g$ being the degree matrix of $\mathcal{W}_g$ with diagonal elements $D_{g,ii} = \sum_j w_{ij}^g$. 

We perform eigen-decomposition $\mathcal{L}_g=U\Lambda {{U}^{\top }}$ to obtain the graph Fourier basis $U$.
Eigen-decomposition of the Laplacian yields a set of eigenvalues and corresponding eigenvectors, where the eigenvalues 
$\Lambda =diag({{\lambda }_{0}},...,{{\lambda }_{N-1}}) $  quantify the associated graph spectral  frequencies and the eigenvectors $ U=[{{u}_{0}};...;{{u}_{N-1}}] $ define orthogonal basis functions over the point cloud domain. We then project the spatial features into the spectral domain via the GFT: 
\begin{align}
	\tilde{X}=GFT(X)={{U}^{\top }}X,\tilde{X}\in {{\mathbb{R}}^{N\times D}}
\end{align}

To enforce coherence, we design a spectral gating network (SGNet) to recalibrate channel-wise responses. We divide channels into groups and learn a frequency-aware modulation weight $W_{spec}$:
\begin{align}
	W_{spec}=SGNet(\tilde{X})=\sigma ({{L}_{1}}(SiLU(LN({{L}_{2}}(\tilde{X})))))
\end{align}
where $L_1$, $L_2$  are learnable linear projections.

This gating selectively emphasizes spectral components that consistently contribute to semantic identity. Finally, project the spectral signals back to the spatial domain by Inverse Graph Fourier Transform (IGFT) to forming the spectral attention map independent of the sequence order:
\begin{align}
	{{A}_{spec}^L}=IGFT({{W}_{spec}})=U{{W}_{spec}}
\end{align}

This instantiation provides an explicit precise realization of global spectral anchoring by directly operating on the Laplacian eigenbasis. It is particularly suitable for moderate-scale point clouds, where exact frequency-domain interpretation is desired.

\textit{SCR-C: Scalable Spectral Anchoring via Chebyshev Polynomials.}
For large-scale point clouds, SCR-C leverages a linear-complexity spectral filter based on Chebyshev polynomials to circumvent the computationally expensive eigendecomposition. As shown in Fig.\ref{architecture}e, like GSC, we construct a KNN graph $\mathcal{G}_c = (\mathcal{V}_c, \mathcal{E}_c, \mathcal{W}_c)$ with $K_S$ neighbors based on the  feature $X$ and the corresponding points set $\mathcal{P}_C$. To ensure numerical stability and well-defined spectral properties, we adopt random walk normalization for the adjacency matrix. Let $\mathcal{W}_c$ denote the weighted adjacency matrix and $D_c$ be the degree matrix. The normalized transition matrix is defined as $D_c^{-1} \mathcal{W}_c$, where each entry ${w}_{ij}^c = w_{ij}^c / \sum_{m \in \mathcal{N}(i)} w_{im}$ represents the normalized edge weight. Consequently, the random walk Graph Laplacian operator is formulated as $\mathcal{L}_c = I -{{D}_c^{-1}}\mathcal{W}_c$.

In implementation, the term  ${{D}_c^{-1}}\mathcal{W}_cX$ in ${\mathcal{L}}_cX$ is efficiently computed via message passing on the sparse neighbor list, avoiding the materialization of dense matrices.

Then, We center the feature map to suppress the zero-frequency component, yielding $X'=X-Mean(X)$. The  $k$-th order Chebyshev polynomial  ${{T}_{k}}(X')$ is computed recursively:
\begin{align}
	& {{T}_{0}}(X')=X', \quad {{T}_{1}}(X')={\mathcal{L}'_c}X', \\ 
	& {{T}_{k}}(X')=2{\mathcal{L}'_c}{{T}_{k-1}}(X')-{{T}_{k-2}}(X'), k\ge2 
\end{align}
where  ${\mathcal{L}'_c}={}^{2{{L}_c}}\!\!\diagup\!\!{}_{{{\lambda }_{max}}}\;-I$ is the rescaled Laplacian for spectral approximation, ${{\lambda }_{max}} = 2.0$ is the upper bound of the eigenvalues.

To explicitly capture multi-scale semantics, we use the three Chebyshev terms ($T_0$, $T_1$, $T_2$) to separate the signal into Low ($F_L$), Mid ($F_M$), and High ($F_H$) frequency bands:
\begin{align}
	F_L = T_0 + T_1, \quad F_M = T_1, \quad F_H = T_2 - T_1
\end{align}

This decomposition isolates global structural smoothness ($F_L$) from sharp geometric variations ($F_H$). Each band is processed by the SGNet, generating band specific modulation masks. The final global attention mask $A_{spec}$ is a learnable weighted sum:
\begin{align}
	A_{spec}^C=\sum\nolimits_{b}{{{\alpha }_{b}}\cdot SGNet({{F}_{b}})}, b\in \{L,M,H\}
\end{align}
where $\alpha =({{\alpha }_{L}},{{\alpha }_{M}},{{\alpha }_{H}})$  represents learnable band importance weights, and $\sum\nolimits_{b}{{{\alpha }_{b}}=1}$.

This instantiation approximates spectral anchoring via Chebyshev polynomials with linear complexity, preserving frequency selectivity while avoiding explicit eigendecomposition, and is therefore suitable for large-scale point cloud understanding. Both SCR-C and SCR-R produce a global anchor that subsequently guides channel recalibration.

\subsubsection{Spatial-Spectral Fusion and SMB Integration} 
While the spectral path provides global semantic anchoring, local spatial continuity remains essential for fine‑grained detail. Therefore, we introduce a parallel spatial context path (see Fig.\ref{architecture}d) that operates directly on the token sequence. Depthwise convolutions aggregate information from neighboring tokens, and further apply a channel shuffle operation to enhance inter-group information exchange:
\begin{align}
	{{X}_{spa}}=Shuffle\left(DWCon{{v}_{3\times 3}}(X)\right)
\end{align}

This path ensures that the refined features maintain local geometric continuity, and complementing the global semantics captured in the spectral domain.

The final calibrated feature $X_{out}$ is obtained by modulating the spatial content with the global spectral attention anchor:
\begin{align}
	{{X}_{out}}=Norm({{X}_{spa}}\odot A_{spec}^{n}+X), n \in \{L,C\}
\end{align}
where $A_{spec}^L$ for SCR-L and $A_{spec}^C$ for SCR-C.

This interaction ensures that local features are validated against the global geometric spectrum, reinforcing semantic coherence across channels. The SCR effectively rectifies the feature distribution before SSM recurrent updates, thereby reversing the semantic dilution in deep spread.

Finally, we integrate the calibrated features into Mamba block. Following recent studies, the original causal convolution is replaced with depthwise convolution to better preserve spatial fidelity. To mitigate the limitations of unidirectional state propagation, we employ a bidirectional SSM that combines the F‑SSM with C-SSM, allowing the model to capture semantic dependencies from both directions.

In summary, SM3D proposes a spectral-aware compensation framework. GSC mitigates the intrinsic low-pass bias of SSMs by injecting geometry-guided high frequency components, while SCR globally recalibrates channel responses to ensure semantic consistency. Within SCR, spectral anchoring is realized through exact (SM3D-L) and approximate (SM3D-C) variants, offering a precision–efficiency trade-off.


\section{Experiments}
\label{Experiments}

\begin{table*}[!ht]
	\centering
	\caption{ Classification on ModelNet40 and ScanObjectNN. We report overall accuracy(\%), number of parameters(\#P), and FLOPs (\#F). We use \textit{rotation} and \textit{scale\&translate} as data augmentation for ScanObjectNN and ModelNet40, respectively.
	}
	\begin{tabular}{rlccccccc}
		\toprule
		        \multirow{2}{*}{ Reference} & \multirow{2}{*}{Methods}        & \multicolumn{3}{c}{ScanObjectNN}  & \multicolumn{2}{c}{ModelNet40 1k P} & \multirow{2}{*}{\#P (M)} & \multirow{2}{*}{\#F (G)} \\
		\cmidrule(l){3-5} \cmidrule(l){6-7} &                                 & OBJ\_BG & OBJ\_ONLY & PB\_T50\_RS & w/o Vot &          w/ Vot           &            ~             &            ~             \\ \hline
		                                                                        \multicolumn{9}{c}{\textit{Supervised Learning Only}}                                                                         \\ \hline
		                            CVPR 17 & PointNet \cite{pointnet}        &  73.3   &   79.2    &    68.0     &  89.2   &             -             &           3.5            &           0.5            \\
		                         NeurIPS 17 & PointNet++   \cite{PointNet2}   &  82.3   &   84.3    &    77.9     &  90.7   &             -             &           1.5            &           1.7            \\
		                             TOG 19 & DGCNN  \cite{dgcnn}             &  82.8   &   86.2    &    78.1     &  92.9   &             -             &           1.8            &           2.4            \\ 
		                            NeurIPS 22 & PointNeXt \cite{PointNeXt}                       &    -    &     -     &    87.7     &  92.9   &             -              &           1.4           &           3.6           \\
		                             JAS 23 & PointConT  \cite{pointcont}     &    -    &     -     &    90.30    &  93.5   &             -             &            -             &            -             \\
		                         NeurIPS 24 & PointMamba \cite{PointMamba}    &  88.30  &   87.78   &    82.48    &    -    &             -             &           12.3           &           3.6            \\
		                          ACM MM 24 & Mamba3D   \cite{Mamba3D}        &  92.94  &   92.08   &    91.81    &  93.4   &             -             &           16.9           &           3.9            \\
		                          ACM MM 25 & HydraMamba  \cite{HydraMamba}   &    -    &     -     &    88.3     &  94.0   &             -             &            -             &            -             \\
		                          AAAI   25 & PCM  \cite{PCM}                 &    -    &     -     &    88.1     &  93.4   &             -             &           34.2           &           45.0           \\
		                          CVPR   25 & SAST  \cite{SAST}               &  92.25  &   91.56   &    87.30    &  92.7   &             -             &           12.3           &           3.6            \\
		                                    & SM3D-L                          &  93.72  &   92.35   &    91.11    &  94.4   &             -             &          17.18           &           4.4            \\ \hline
		                                                                   \multicolumn{9}{c}{\textit{With Self-supervised Pre-training}}                                                                     \\ \hline
		                            CVPR 22 & Point-BERT \cite{Point-BERT}    &  87.43  &   88.12   &    83.07    &  92.7   &           93.2            &           23.8           &           4.8            \\
		                            ECCV 22 & Point-MAE \cite{point-mae}      &  92.77  &   91.22   &    89.04    &  92.7   &           93.8            &           23.8           &           4.8            \\
		                         NeurIPS 23 & PointGPT-S   \cite{Pointgpt}    &  93.39  &   92.43   &    89.17    &  93.3   &           94.0            &           29.2           &           5.7            \\
		                            AAAI 24 & Point-FEMAE \cite{point-femae}  &  95.18  &   93.29   &    90.22    &  94.0   &           94.5            &           27.4           &           3.6            \\
		                         NeurIPS 24 & PointMamba  \cite{PointMamba}   &  94.32  &   92.60   &    89.31    &  93.6   &           94.1            &           12.3           &           3.6            \\
		                          ACM MM 24 & Mamba3D \cite{Mamba3D}          &  93.12  &   92.08   &    92.05    &  94.7   &           95.1            &           16.9           &           3.9            \\
		                          ICCV   25 & StruMamba3D  \cite{strumamba3d} &  95.18  &   93.63   &    92.75    &  95.1   &           95.4            &           15.8           &           4.0            \\
		                            ICCV 25 & Point-PQAE \cite{point-pqae}    &  95.0   &   93.6    &    89.6     &  94.0   &             -             &           22.1           &            -             \\
		                           CVPR  25 & SAST   \cite{SAST}              &  94.32  &   91.91   &    89.10    &  93.4   &             -             &           12.3           &           3.6            \\
		                                    & SM3D-C                          &  94.72  &   93.21   &    92.37    &  94.9   &           95.4            &           17.0           &           4.1            \\
		                                    & SM3D-L                          &  95.17  &   93.83   &    93.17    &  95.6   &           96.0            &           17.1           &           4.4            \\ \bottomrule
	\end{tabular}
	\label{Class}
\end{table*}

\subsection{Implementation Details}

We implement SM3D using PyTorch on a single NVIDIA TITAN RTX 24GB GPU. 
The backbone hidden dimension $D$ is $384$. The input point cloud is tokenized into $N=128$ patches via FPS, with each patch containing $K=32$ neighbors grouped by KNN. In the GSC and SCR-C, the spectral graph size is set to $K_G=4$ and $K_S=32$.
We employ the AdamW optimizer with an initial learning rate of $5 \times 10^{-4}$ and a weight decay of 0.05. The learning rate follows a cosine decay schedule with a 10-epoch linear warmup. The model is trained for 300 epochs with a batch size of 32. Cross-entropy loss is adopted for supervision. Following established protocols \cite{Point-BERT}, \cite{point-mae}, \cite{PointMamba}, \cite{Mamba3D}, we conduct self-supervised pre-training on the ShapeNetCore \cite{ShapeNet}  dataset before fine-tuning.

\vspace{-2.5mm}

\subsection{Downstream Tasks}

\subsubsection{Object classification}

We evaluate SM3D on ModelNet40 \cite{modelnet} and ScanObjectNN \cite{ScanObjectNN} for 3D object classification. ModelNet40 contains 12,311 CAD models from 40 categories. ScanObjectNN is a real-world dataset comprising approximately 15K objects from 15 indoor categories and includes three variants: OBJ\_BG, OBJ\_ONLY, and the most challenging PB\_T50\_RS (with perturbations). For ModelNet40, 1,024 points are uniformly sampled from each object, while 2,048 points are used for ScanObjectNN. For pre-training, we adopt the masking ratio (60\%) that was used following the previous protocol \cite{Point-BERT}, \cite{point-mae}, \cite{PointMamba}, \cite{SAST}.

\begin{figure*}[!ht]
	\centering
	\includegraphics[width=0.8\textwidth]{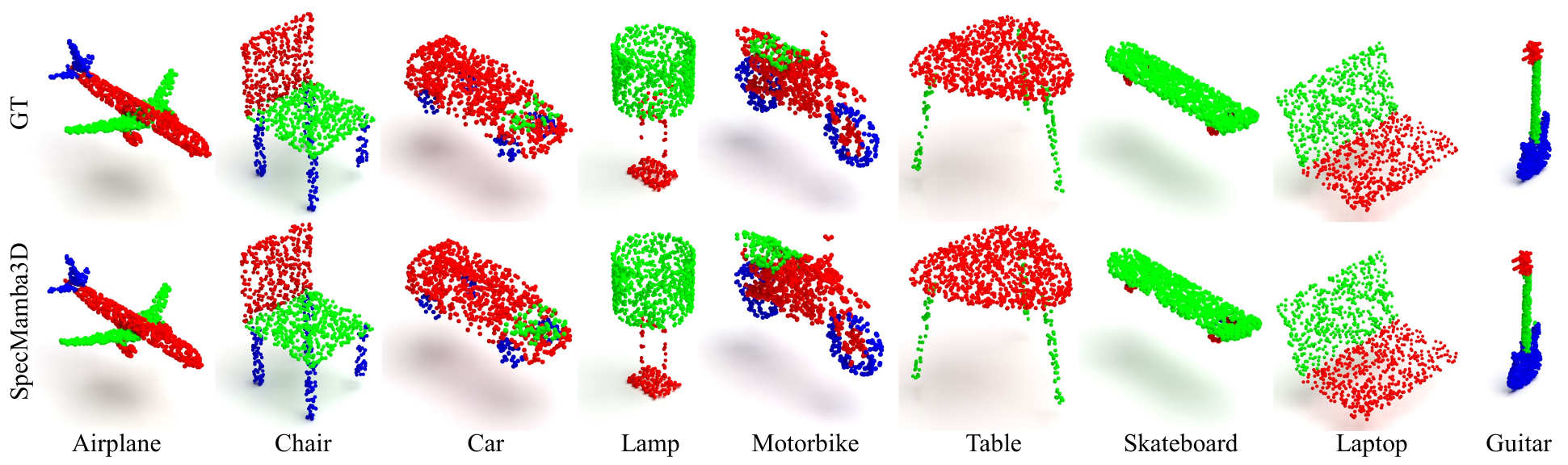}
	\caption{Qualitative results of part segmentation by SM3D-L on ShapeNetPart. The top row is the Ground Truth (GT).}\label{fig partseg}
\end{figure*}

\textit{Result.}
As shown in Table \ref{Class}, SM3D achieves a state-of-the-art results.
Without pretraining, our method significantly outperforms existing Mamba-based variants. Notably, on the challenging PB\_T50\_RS benchmark, SM3D-L achieves 91.11\%, surpassing the baseline PointMamba \cite{PointMamba} by a remarkable margin of 8.63\%. This substantial gain validates that GSC effectively recovers high frequency spectral component from perturbed data. Furthermore, SM3D-L outperforms Mamba3D \cite{Mamba3D} by 0.78\% on OBJ\_BG with comparable parameters (17.18M vs 16.9M), confirming the superiority of our spectral-aware design over purely spatial aggregation.

With self-supervised pretraining, SM3D further improve performance. SM3D-C achieves up to 96.0\% accuracy on ModelNet40 with voting, while SM3D-L attains 93.17\% accuracy on PB\_T50\_RS without voting, surpassing StruMamba3D by 0.42\%. Compared to recent methods such as Point-FEMAE \cite{point-femae} (Transformer-based) and Point-PQAE \cite{point-pqae} (Reconstruction-based), SM3D-L exhibits substantial improvements of 4.15\% and 3.57\% on PB\_T50\_RS, respectively, highlighting its robustness under severe geometric perturbations. These gains are achieved with only 17.1M parameters and 4.4G FLOPs, demonstrating a favorable accuracy–efficiency trade-off.

\subsubsection{Few-shot learning}

We evaluate the few-shot generalization capability of SM3D on the ModelNetFewShot benchmark under the standard $N$-way $K$-shot protocol \cite{Point-BERT}, \cite{Mamba3D}, \cite{PointMamba}, \cite{SAST}.. All results are reported as the mean overall accuracy and standard deviation over 10 independent trials.

\begin{table}[t]
	\centering
	
	\caption{\textbf{Few-shot classification on ModelNet40 dataset.} A$_{\pm std}$ represents the average(A) and standard deviation($std$).
	}
	
		\begin{tabular}{lcccc}
			\toprule
			\multirow{2}{*}{Methods}                  &    \multicolumn{2}{c} {5-Way}     &    \multicolumn{2}{c}{10-Way}     \\
			\cmidrule (l){2-3} \cmidrule(l){4-5}
			~ &     10-Shot     &     20-Shot     &     10-Shot     &     20-Shot     \\ \hline
			\multicolumn{5}{c}{\textit{Supervised Learning Only}}                           \\ \hline
			PointNet \cite{pointnet}                                  & 52.0$_{\pm3.8}$ & 57.8$_{\pm4.9}$ & 46.6$_{\pm4.3}$ & 35.2$_{\pm4.8}$ \\
			DGCNN  \cite{dgcnn}                                   & 31.6$_{\pm2.8}$ & 40.8$_{\pm4.6}$ & 19.9$_{\pm2.1}$ & 16.9$_{\pm1.5}$ \\
			SM3D-L                                    & 91.9$_{\pm3.6}$ & 96.7$_{\pm2.2}$ & 86.5$_{\pm4.8}$ & 92.1$_{\pm3.4}$ \\ \hline
			\multicolumn{5}{c}{\textit{With Self-supervised Pre-training} }                     \\ \hline
			OcCo  \cite{OcCo}                         & 94.0$_{\pm3.6}$ & 95.9$_{\pm2.7}$ & 89.4$_{\pm5.1}$ & 92.4$_{\pm4.6}$ \\
			ACT \cite{ACT}                            & 96.8$_{\pm2.3}$ & 98.0$_{\pm1.4}$ & 93.3$_{\pm4.0}$ & 95.6$_{\pm2.8}$ \\
			MaskPoint \cite{MaskPoint}                & 95.0$_{\pm3.7}$ & 97.2$_{\pm1.7}$ & 91.4$_{\pm4.0}$ & 93.4$_{\pm3.5}$ \\
			Point-BERT \cite{Point-BERT}              & 94.6$_{\pm3.1}$ & 96.3$_{\pm2.7}$ & 91.0$_{\pm5.4}$ & 92.7$_{\pm5.1}$ \\
			Point-MAE \cite{point-mae}                & 96.3$_{\pm2.5}$ & 97.8$_{\pm1.8}$ & 92.6$_{\pm4.1}$ & 95.0$_{\pm3.0}$ \\
			PointGPT-S \cite{Pointgpt}                & 96.8$_{\pm2.0}$ & 98.6$_{\pm1.1}$ & 92.6$_{\pm4.6}$ & 95.2$_{\pm3.4}$ \\
			PointMamba \cite{PointMamba}              & 95.0$_{\pm2.3}$ & 97.3$_{\pm1.8}$ & 91.4$_{\pm4.4}$ & 92.8$_{\pm4.0}$ \\
			SM3D-C                                    & 95.2$_{\pm4.8}$ & 97.4$_{\pm1.8}$ & 91.1$_{\pm4.2}$ & 94.5$_{\pm3.3}$ \\
			SM3D-L                                    & 96.7$_{\pm5.0}$ & 98.7$_{\pm1.2}$ & 92.4$_{\pm5.0}$ & 95.8$_{\pm3.0}$ \\ \bottomrule
		\end{tabular}
		
	
	\label{fewshot}
\end{table}

\textit{Results.} 
Table \ref{fewshot} demonstrates the strong generalization of our framework. In the supervised-only setting, SM3D-L achieves 91.9\% (5-way 10-shot) and 86.5\% (10-way 10-shot), significantly outperforming DGCNN \cite{dgcnn} baselines. This suggests that explicitly modeling high-frequency geometric priors acts as a powerful inductive bias, reducing overfitting when data is scarce.
\begin{table}[!t]
	\centering
	
	\caption{\textbf{Part segmentation on the ShapeNetPart.} We reports class-level mIoU (mIoU$_C$) and instance-level mIoU (mIoU$_I$).}
	
		\begin{tabular}{rlcc}
			\toprule
			Reference & Method                        & mIoU$_C$(\%) & mIoU$_I$(\%) \\ \hline
			\multicolumn{4}{c}{\textit{Supervised Learning Only}}           \\ \hline
			CVPR 17 & PointNet \cite{pointnet}      &     80.4     &     83.7     \\
			NeurIPS 17 & PointNet++ \cite{PointNet2}   &     81.9     &     85.1     \\
			TOG 19 & DGCNN \cite{dgcnn}            &     82.3     &     85.2     \\
			CVPR 23 & APES   \cite{APES}                &     83.7     &     85.8     \\ \hline
			\multicolumn{4}{c}{\textit{With Self-supervised Pre-training}}      \\ \hline
			ICCV 21 & OcCo   \cite{OcCo}            &     83.4     &     84.7     \\
			ECCV 22 & MaskPoint \cite{MaskPoint}    &     84.7     &     86.2     \\
			ECCV 22 & Point-MAE \cite{point-mae}    &     84.2     &     86.1     \\
			CVPR 22 & Point-BERT \cite{Point-BERT}  &     84.1     &     85.6     \\
			NeurIPS 23 & PointGPT-S \cite{Pointgpt}    &     84.1     &     86.2     \\
			ICLR 23 & ACT       \cite{ACT}          &     84.7     &     86.2     \\
			NeurIPS 24 & PointMamba  \cite{PointMamba} &     84.4     &     86.0     \\
			ACM MM 24 & Mamba3D   \cite{Mamba3D}      &     83.6     &     85.6     \\
			ICCV 25 & Point-PQAE  \cite{point-pqae} &     84.6     &     86.1     \\
			AAAI 25 & PCM-T      \cite{PCM}         &     84.6     &     86.0     \\
			& SM3D-C                        &     84.8     &     86.3     \\
			& SM3D-L                        &     84.9     &     86.5     \\ \bottomrule
		\end{tabular}
	
	\label{tab partseg}
\end{table}

With pre-training, both SM3D variants achieve competitive performance against recent methods. SM3D-L's accuracy match or slightly surpass PointGPT-S \cite{Pointgpt} and Point-MAE \cite{point-mae} on the 5-way tasks, while maintaining comparable variance. Comparing the two variants, SM3D-L consistently outperforms SM3D-C by margins of 0.7\%-1.5\% and exhibits lower standard deviation. This indicates that exact spectral decomposition provides more precise and stable semantics anchors for few-shot discrimination. However, as the sample size increases, the gap narrows, validating that the Chebyshev approximation effectively captures the dominant spectral features required for generalization. These results demonstrate that spectral modeling facilitates robust few-shot learning.

\subsubsection{Part segmentation}

We evaluate fine-grained part segmentation on the ShapeNetPart \cite{ShapeNet}, which contains approximately 16K shapes from 16 object categories with 50 part labels. Following standard protocols \cite{PointMamba}, \cite{Mamba3D}, \cite{SAST}, 2,048 points are uniformly sampled from each shape without using normals. Quantitative results are reported in Table \ref{tab partseg}, qualitative visualizations are provided in Fig.\ref{fig partseg}.

\textit{Results.}
ShapeNetPart \cite{ShapeNet} is a highly challenging benchmark, with recent methods exhibiting only incremental improvements. Nevertheless, SM3D-L achieves 86.5\% Ins.mIoU, outperforming PointMamba \cite{PointMamba} and Mamba3D \cite{Mamba3D} by 0.5\% and 0.9\%, respectively. Both SM3D variants consistently surpass Transformer-based methods such as Point-MAE \cite{point-mae} and PointGPT-S \cite{Pointgpt}. Notably, the performance gap between SM3D-L and SM3D-C is minimal (0.1\% Cls.mIoU, 0.2\% Ins.mIoU). This suggests that for dense segmentation tasks, the polynomial spectral approximation is sufficient to capture the required local-global consistency, offering a favorable accuracy-efficiency trade-off. Qualitatively, our method excels at delineating boundaries in complex structures such as airplane engines, motorbike wheels (see Fig.\ref{fig partseg}). This sharpness is a direct benefit of the GSC module, which preserves high frequency edge information. It should be noted that the car hood and motorbike seat are slightly different from the GT, which means that the model still needs further improvement.

\subsubsection{Scene-level semantic segmentation}

We evaluate scene-level semantic segmentation on the S3DIS \cite{s3dis}. Following the standard protocol \cite{ptv3}, \cite{PCM}, \cite{Pamba}, Area 5 is used for testing, and the remaining areas are used for training. Due to the massive scale of point clouds, we employ the SM3D-C for this task. As shown in Table \ref{semseg}, SM3D-C achieves 66.7\% mIoU and 71.3\% mAcc, outperforming recent models such as Point-PQAE \cite{point-pqae} and PCM \cite{PCM} by 5.3\% and 3.3\% in mIoU. These improvements indicate that spectral compensation remains effective for large-scale, cluttered indoor scenes. We note that methods such as PTv3 \cite{ptv3}, SpoTr \cite{spotr}, and Pamba \cite{PCM} achieve higher absolute performance. These approaches are explicitly designed for scene-level semantic segmentation and incorporate task-specific optimizations. In contrast, SM3D-C prioritizes generality while maintaining competitive performance across diverse tasks.

\begin{table}
	\centering
	\caption{\textbf{3D semantic segmentation on the S3DIS.} The mean accuracy (mAcc) and mean IoU (mIoU) are reported.}
	\centering
	\resizebox{0.49\textwidth}{!}{
		\begin{tabular}{rlcc}
			\toprule
			Reference & Method             & mAcc (\%) & mIoU (\%) \\ \midrule
			CVPR 17 & PointNet \cite{pointnet}   &   49.0    &   41.1    \\
			NeurIPS 17 & PointNet++ \cite{PointNet2} &   67.1    &   53.5    \\ 
			ECCV 22 & Point-MAE \cite{point-mae}  &   69.9    &   60.8    \\ 
			NeurIPS 23 & PointGPT-L  \cite{Pointgpt}       &   70.6    &   62.2    \\
			ICLR 23 & ACT    \cite{ACT}            &   71.1    &   61.2    \\ 
			CVPR 23 & SpoTr     \cite{spotr}         &   76.4    &   70.8    \\
			CVPR 24 & PTv3    \cite{ptv3}           &   78.9    &   73.4    \\
			CVM 25 & Siwn3D    \cite{swin3d}         &     -     &   72.5    \\
			ICCV 25 & Point-PQAE   \cite{point-pqae}      &   70.6    &   61.4    \\ 
			AAAI 25 & PCM       \cite{PCM}         &     -     &   63.4    \\
			AAAI 25 & Pamba     \cite{Pamba}         &     -     &   73.5    \\
			& SM3D-C             &   71.3    &   66.7    \\ \bottomrule
		\end{tabular}
	}
	\label{semseg}
\end{table}

\begin{figure}[!t] 
	\centering
	\includegraphics[width=1.65in]{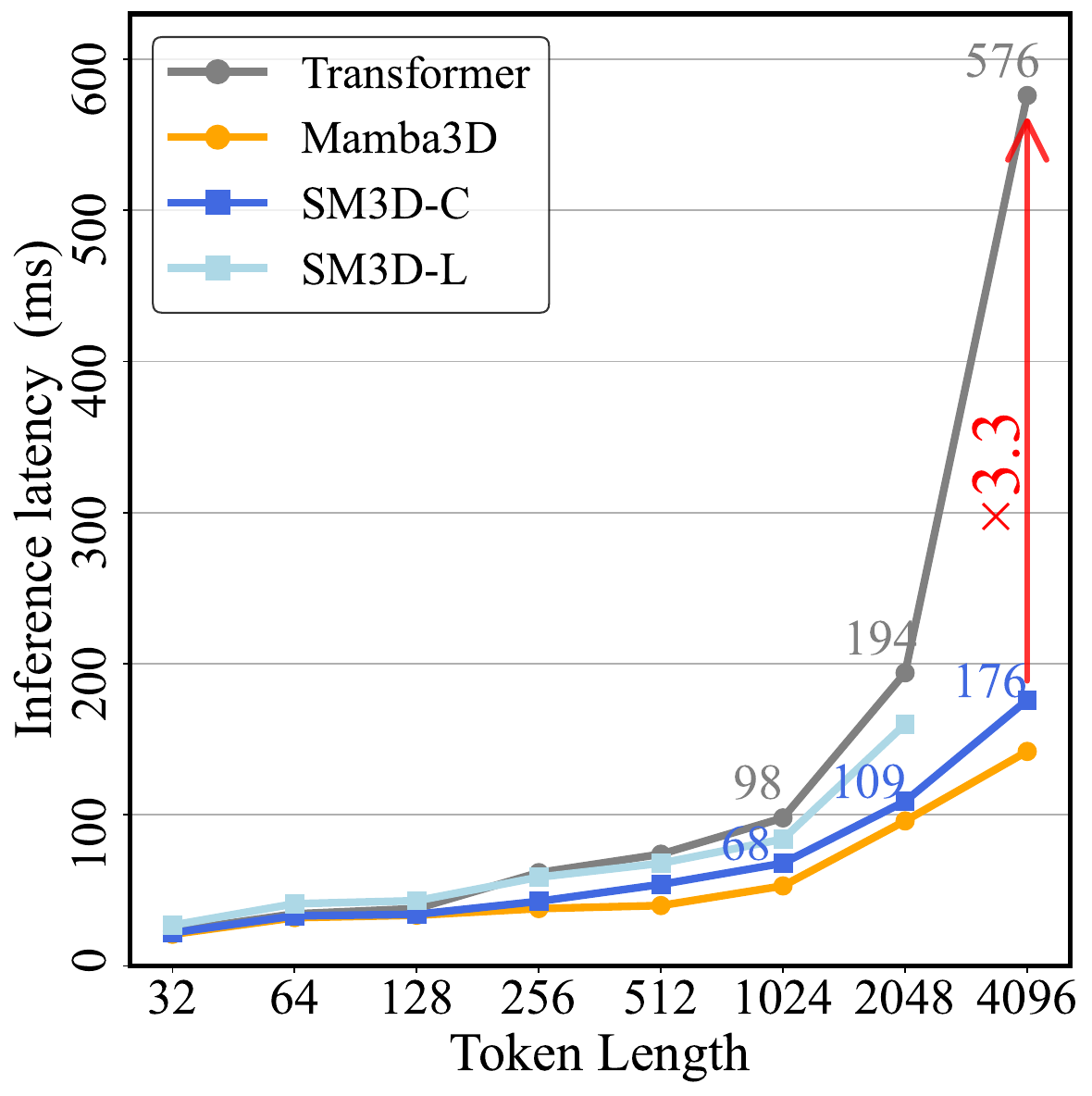}
		\label{fig_time}
	\hfil
	\includegraphics[width=1.65in]{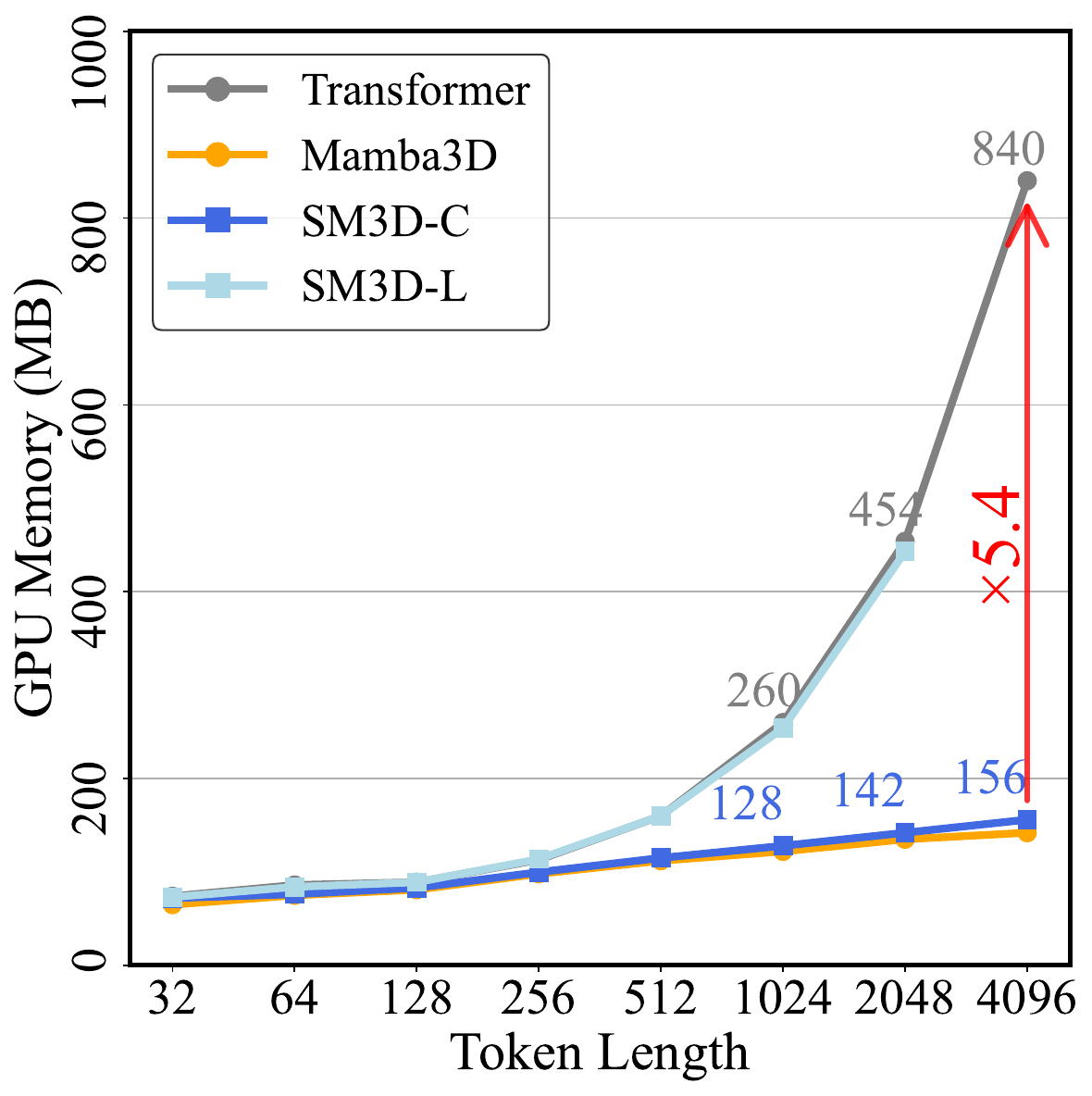}
		\label{fig_gpu}
	\caption{Computational efficiency analysis of SM3D. Left: Inference time on NVIDIA TiTan RTX. Right: GPU memory usage for per sample.} 
	\label{fig_cost} 
\end{figure}

\subsubsection{Efficiency Analysis}

 We analyze the theoretical complexity and real-world resource consumption of SM3D with a batch size of 1.
For SM3D-L, the graph Laplacian eigendecomposition is performed once as preprocessing to avoiding repeated, reducing calculate to $\mathcal{O}(N^2)$. However, eigendecomposition becomes numerically unstable beyond 2,048 tokens due to repeated eigenvalues, directly motivating the design of SM3D-C. SM3D-C employs low-order Chebyshev polynomial approximation without dense $N \times N$ matrices on sparse graphs, achieving linear complexity with respect to the number of edges.
As shown in Fig.\ref{fig_cost}, SM3D-C exhibits a linear growth curve for GPU memory and inference latency comparable to Mamba3D \cite{Mamba3D} and significantly lower than Transformers and SM3D-L, confirms that SM3D-C is the balanced choice for large-scale scene understanding.

\vspace{-2mm}

\subsection{Ablation study}

\subsubsection{The effect of each component }
We analyze the contribution of the GSC and the SCR by progressively integrating them into the baseline BiSSM backbone.  Results are summarized in Table \ref{tab module-level ablation}.

\begin{table}[t]
	\centering
	
	\caption{ Component-level ablation on ScanObjectNN.}
		
		\begin{tabular}{ccccccc}
			\toprule
			BiSSM   &    GSC    &    SCR    & OBJ\_ONLY & PB\_T50\_RS & \#P (M) & \#F (G) \\ \hline
			\ding{51} & \ding{55} & \ding{55} &   89.51   &    85.32    &  13.34  &  3.60   \\
			\ding{51} & \ding{51} & \ding{55} &   91.22   &    89.67    &  16.95  &  4.08   \\
			\ding{51} & \ding{55} & \ding{51} &   90.98   &    88.32    &  13.54  &  3.94   \\
			\ding{51} & \ding{51} & \ding{51} &   92.35   &    91.11    &  17.18  &  4.42   \\ \bottomrule
	\end{tabular}	
	\label{tab module-level ablation}
\end{table}

\textit{Impact of GSC.} 
Introducing the GSC module leads to a substantial performance improvement. Specifically, accuracy increases 1.71\% and 4.35\% on OBJ\_ONLY and PB\_T50\_RS. This pronounced gain on the perturbed benchmark validates standard SSMs intrinsically suppress high-frequency signals, leading to vulnerability under geometric perturbations. By explicitly injecting graph-guided high-frequency residuals, GSC effectively counteracts this low-pass bias, allowing the model to recover discriminative structural details.

\textit{Impact of SCR.} 
The addition of SCR independently improves the baseline by 1.47\% on OBJ\_ONLY and 3.00\% on PB\_T50\_RS. While the numerical gain is slightly lower than that of GSC, deep recurrence in SSMs leads to semantic dilution. SCR mitigates this issue by anchoring spatial features in a global spectral space and performing frequency-aware channel recalibration, thereby stabilizing semantic representations during deep propagation. From an efficiency perspective, the full model only introduces an additional 3.84M parameters and 0.82G FLOPs compared to the baseline.

\subsubsection{Ablation study on the design choices}

We validate the rationality of our architectural design through fine-grained ablations on internal sub-components and spectral operators.

\textit{Impact of Internal Sub-components.} 
For GSC (see Table \ref{ablation on internal components}a), the most critical drop occurs when removing the Laplacian Operator, confirming that explicit high-frequency extraction is the core driver of performance. Removing the normalization and Affine Transform also degrades accuracy, as canonicalizing local patches is importation for invariant spectral analysis.
For SCR (see Table \ref{ablation on internal components}b), the Spectral Path proves indispensable, with its removal causing a 2.66\% drop on PB\_T50\_RS. This validates that purely spatial convolution fails to maintain semantic coherence under perturbation. The SGNet and Channel Shuffle operations provide further gains by enhancing non-linear feature mixing.

\begin{table}[!t]
	\centering
	\caption{Fine-grained ablation on internal components.}
	\resizebox{0.5\textwidth}{!}{
		\begin{tabular}{lcccc}
			\toprule
			Component Config.                                      & OBJ\_ONLY      & PB\_T50\_RS    & \#P (M) & \#F (G) \\ \hline
			\multicolumn{5}{l}{\textit{(a) Ablation on GSC components.}}                                           \\
			$\star$ \textbf{Full Model}                            & \textbf{92.35} & \textbf{91.11} & 17.18   & 4.42    \\
			\hspace{1em} w/o Norm. and  Affine & 91.75          & 90.74          & 17.17   & 4.42    \\
			\hspace{1em} w/o Laplacian Operator                    & 90.13          & 89.35          & 17.18   & 4.00    \\
			\hspace{1em} w/o Gaussian Weighting                    & 91.03          & 90.45          & 17.18   & 4.42    \\
			\multicolumn{5}{l}{\textit{(b) Ablation on SCR components.}}                                           \\
						\hspace{1em} w/o Spectral Path                         & 91.05          & 88.45          & 17.18   & 4.12    \\
			\hspace{1em} w/o SGNet                                 & 91.78          & 90.21          & 17.18   & 4.42    \\
			\hspace{1em} w/o Conv                                  & 91.60          & 89.33          & 17.01   & 4.40    \\
			\hspace{1em} w/o Channel Shuffle                       & 91.53          & 89.12          & 17.18   & 4.42    \\ \bottomrule
		\end{tabular}
	}
	\label{ablation on internal components}
\end{table}

\begin{table}[!t]
	\centering
	\caption{Ablation on Spectral Operators and Basis Variants.}
	\resizebox{0.48\textwidth}{!}{
		\begin{tabular}{llcc}
			\toprule
			Operator/Basis Variant      & Definition                  & OBJ\_ONLY      & PB\_T50\_RS    \\ \midrule
			\multicolumn{4}{l}{\textit{(a) Spectral operator of GSC.}}                                  \\
			Identity                    & $I$                         & 90.20          & 89.26          \\
			Low-pass Smoothing          & $D^{-1}W$                   & 89.61          & 88.57          \\
			\textbf{Graph Laplacian (Ours)}    & Eq. \eqref{L}      & \textbf{92.35} & \textbf{91.11} \\ \midrule
			\multicolumn{4}{l}{\textit{(b) Spectral Paths  in SCR.}}                              \\
			Random Basis                & $U \to U_{Random}$                   & 90.43          & 88.46          \\
			Low-freq Basis              & $U \to U_{[1:N/3]}$         & 90.95          & 89.62          \\
			\textbf{Full-freq Basis (Ours)}    & $U_{[1:N]}$        & \textbf{92.35} & \textbf{91.11} \\ \bottomrule
		\end{tabular}
	}
	\label{Ablation on different variant choices}
\end{table}

\textit{Analysis of Spectral Operators and Bases.} 
In Table \ref{Ablation on different variant choices}, we investigate the choice of spectral formulation.
For GSC (see Table \ref{Ablation on different variant choices}a), a key observation is that Low-pass Smoothing performs worse than the Identity mapping. This aligns with our theoretical analysis in Section \ref{problem}: since the SSM backbone already acts as a low-pass filter, adding extra smoothing exacerbates feature blurring. In contrast, our Graph Laplacian explicitly acts as a high-pass filter, effectively compensating for the missing high-frequency details.
For SCR (see Table \ref{Ablation on different variant choices}b), using a Full Frequency Basis significantly outperforms restricted Low-frequency or Random bases. This suggests that semantic consistency is not just about global shape (low-freq) but also relies on fine-grained spectral signatures (high-freq) to distinguish subtle category differences.

\vspace{-2mm}
\subsection{Heatmaps Visualization and Spectral Evolution}

\begin{figure}[!tbp]
	\centering
	\begin{minipage}[b]{0.15\linewidth}
		\includegraphics[width=\linewidth]{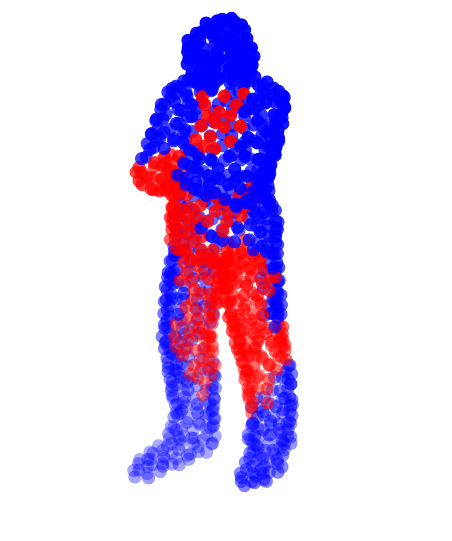}
	\end{minipage}
	\hfill
	\begin{minipage}[b]{0.17\linewidth}
		\includegraphics[width=\linewidth]{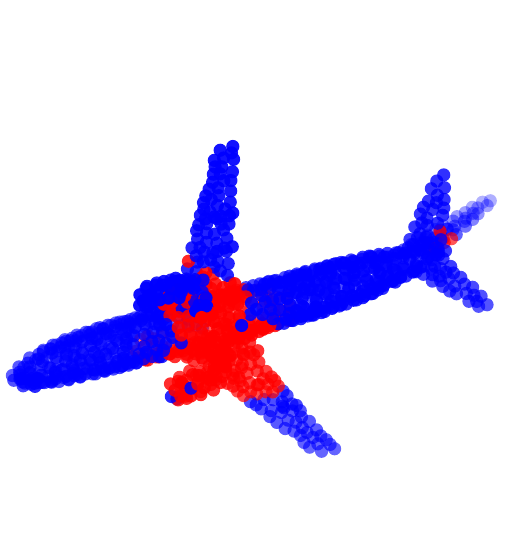}
	\end{minipage}
	\hfill
	\begin{minipage}[b]{0.15\linewidth}
		\includegraphics[width=\linewidth]{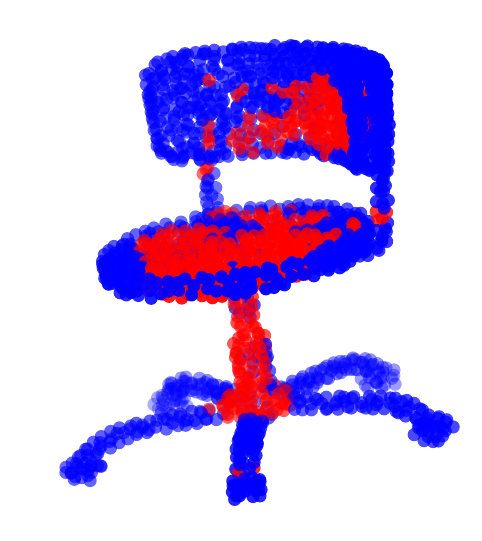}
	\end{minipage}
	\hfill
	\begin{minipage}[b]{0.15\linewidth}
		\includegraphics[width=\linewidth]{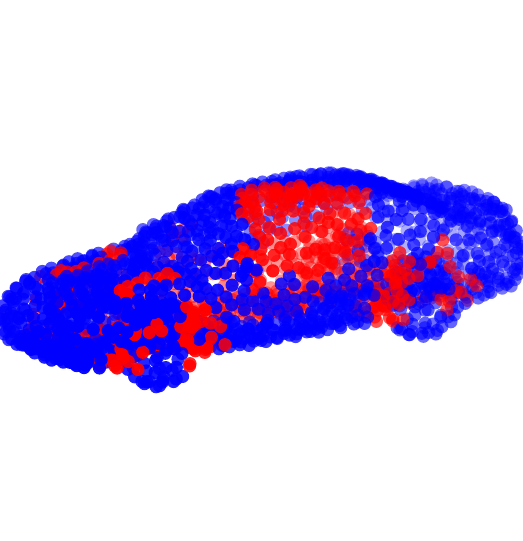}
	\end{minipage}
	\hfill
	\begin{minipage}[t]{0.15\linewidth}
		\includegraphics[width=\linewidth]{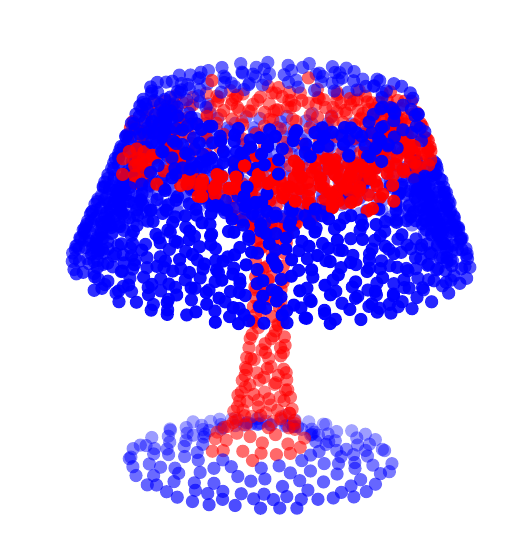}
	\end{minipage}
	\hfill
	\begin{minipage}[t]{0.15\linewidth}
		\includegraphics[width=\linewidth]{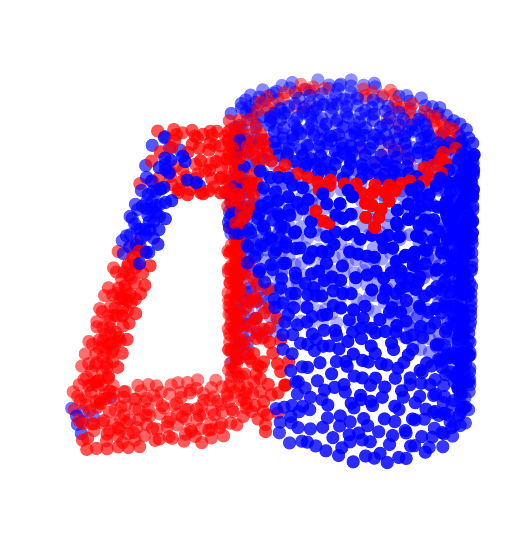}
	\end{minipage}
	
	\begin{minipage}[t]{0.15\linewidth}
		\includegraphics[width=\linewidth]{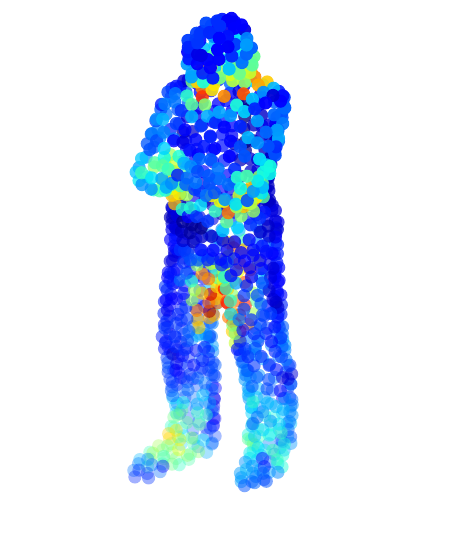}
	\end{minipage}
	\hfill
	\begin{minipage}[t]{0.18\linewidth}
		\includegraphics[width=\linewidth]{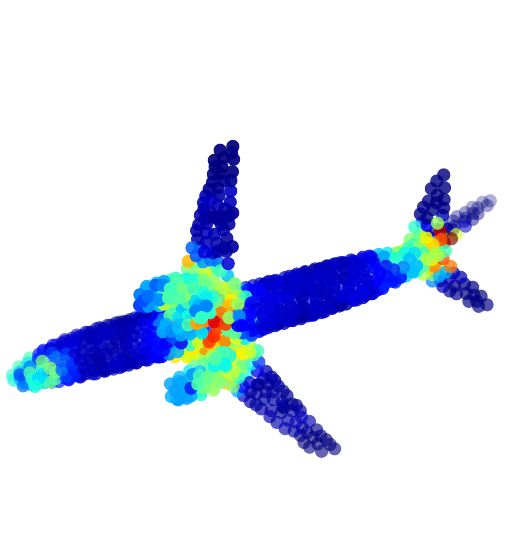}
	\end{minipage}
	\hfill
	\begin{minipage}[t]{0.15\linewidth}
		\includegraphics[width=\linewidth]{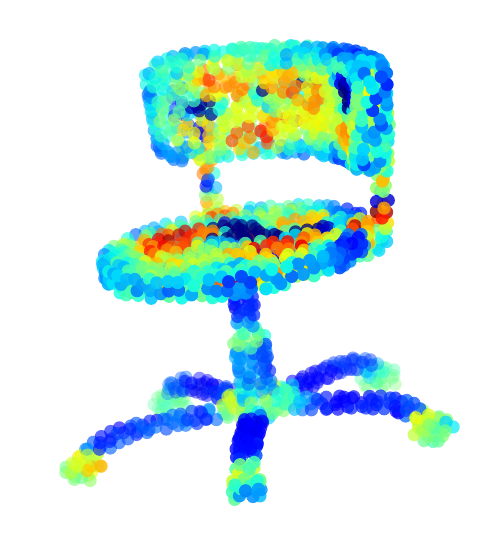}
	\end{minipage}
	\hfill
	\begin{minipage}[t]{0.15\linewidth}
		\includegraphics[width=\linewidth]{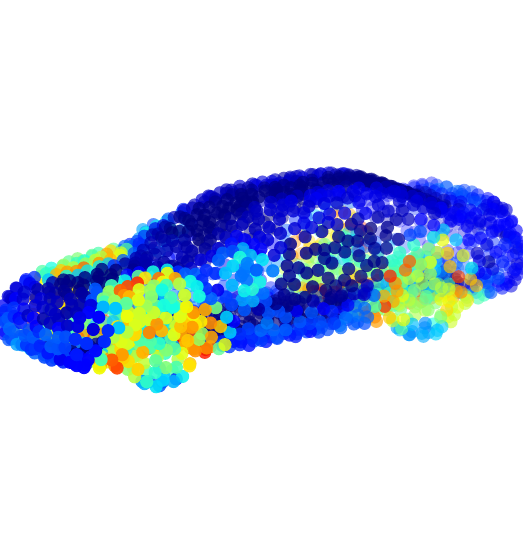}
	\end{minipage}
	\hfill
	\begin{minipage}[t]{0.15\linewidth}
		\includegraphics[width=\linewidth]{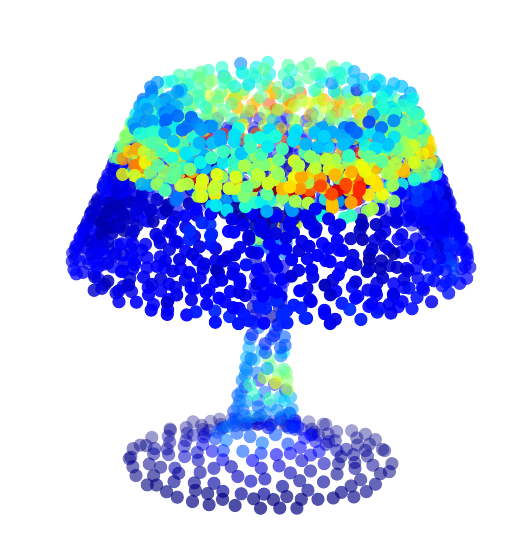}
	\end{minipage}
	\hfill
	\begin{minipage}[t]{0.15\linewidth}
		\includegraphics[width=\linewidth]{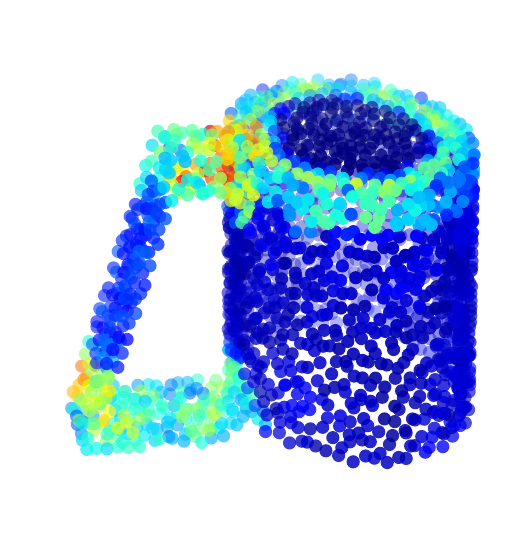}
	\end{minipage}
	
	\begin{minipage}[t]{0.15\linewidth}
		\includegraphics[width=\linewidth]{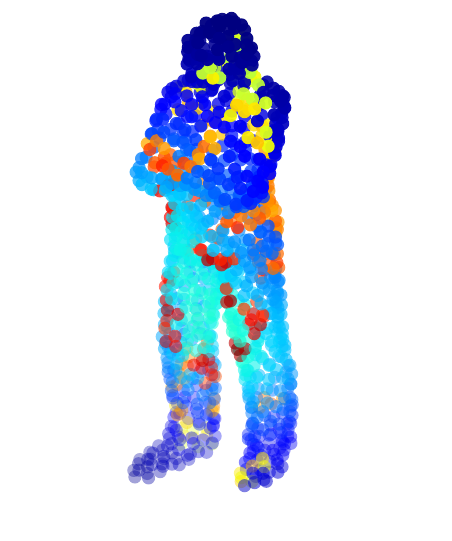}
	\end{minipage}
	\hfill
	\begin{minipage}[t]{0.18\linewidth}
		\includegraphics[width=\linewidth]{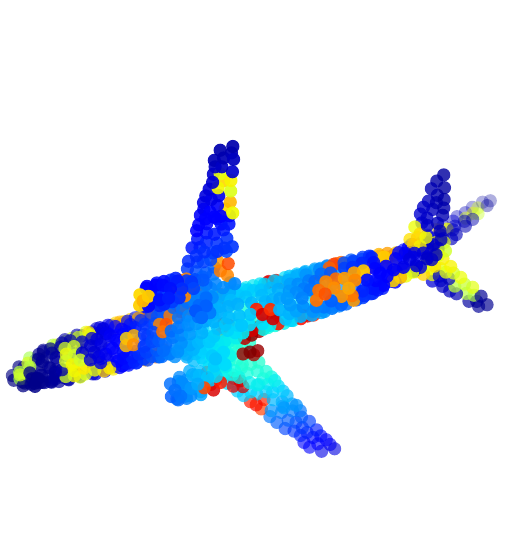}
	\end{minipage}
	\hfill
	\begin{minipage}[t]{0.15\linewidth}
		\includegraphics[width=\linewidth]{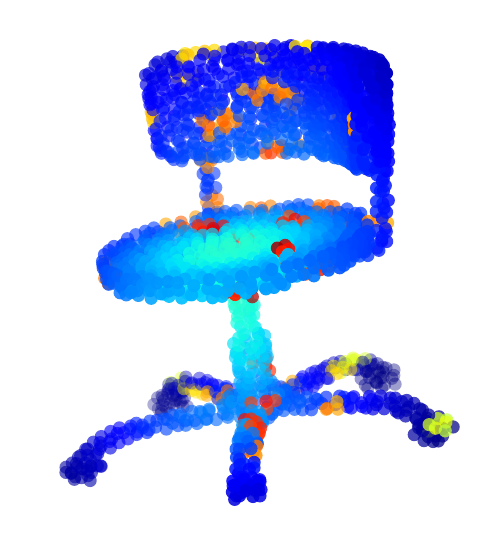}
	\end{minipage}
	\hfill
	\begin{minipage}[t]{0.15\linewidth}
		\includegraphics[width=\linewidth]{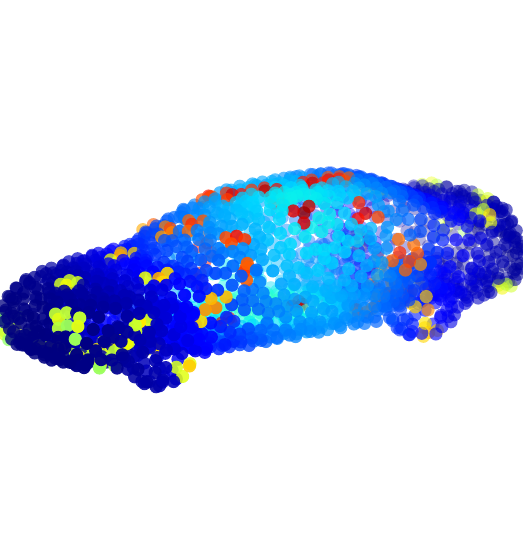}
	\end{minipage}
	\hfill
	\begin{minipage}[t]{0.15\linewidth}
		\includegraphics[width=\linewidth]{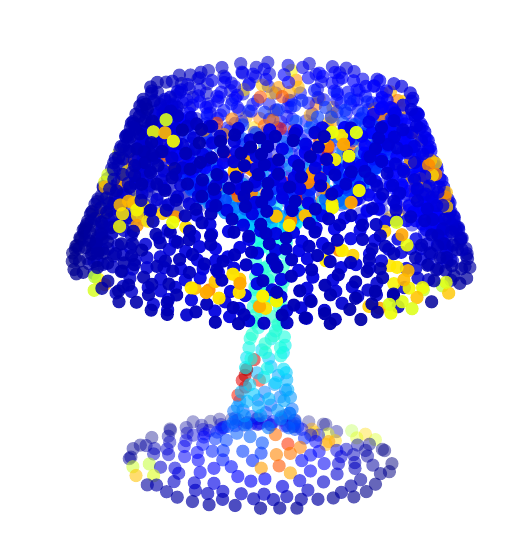}
	\end{minipage}
	\hfill
	\begin{minipage}[t]{0.15\linewidth}
		\includegraphics[width=\linewidth]{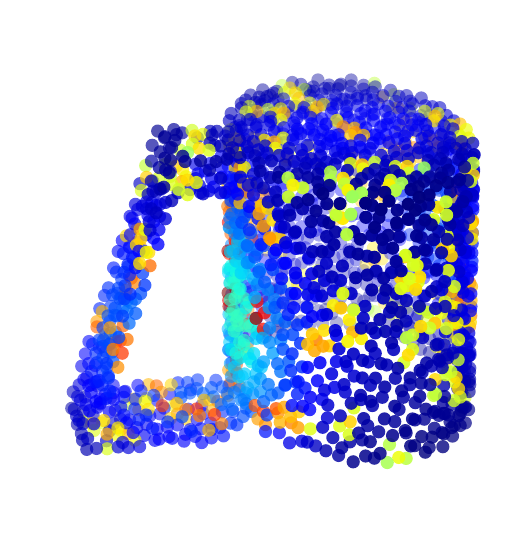}
	\end{minipage}
	
	\caption{
		Qualitative visualizations on ModelNet40. Row 1: Grad-CAM activation maps. Row 2: High frequency spectral energy maps. Row 3: GSC geometric weights.
	}
	\label{heatmap}
\end{figure}

\textit{Visual Interpretability.}
As shown in Fig.\ref{heatmap}, Grad-CAM \cite{grad-cam} activations heavily focus on structurally sharp regions, such as airplane engines, chair legs, and cup handles. Crucially, these regions spatially align with the high-energy areas in the Spectral and Geometric Weight maps. This alignment empirically proves the GSC module successfully amplifies high-frequency geometric residuals, and the network explicitly utilizes these compensated high frequency cues for semantic decision-making. This confirms that our spectral-aware design effectively redirects the model's attention to fine-grained structural details.

\begin{figure}[!t] 
	\centering
	{\includegraphics[width=1.6in]{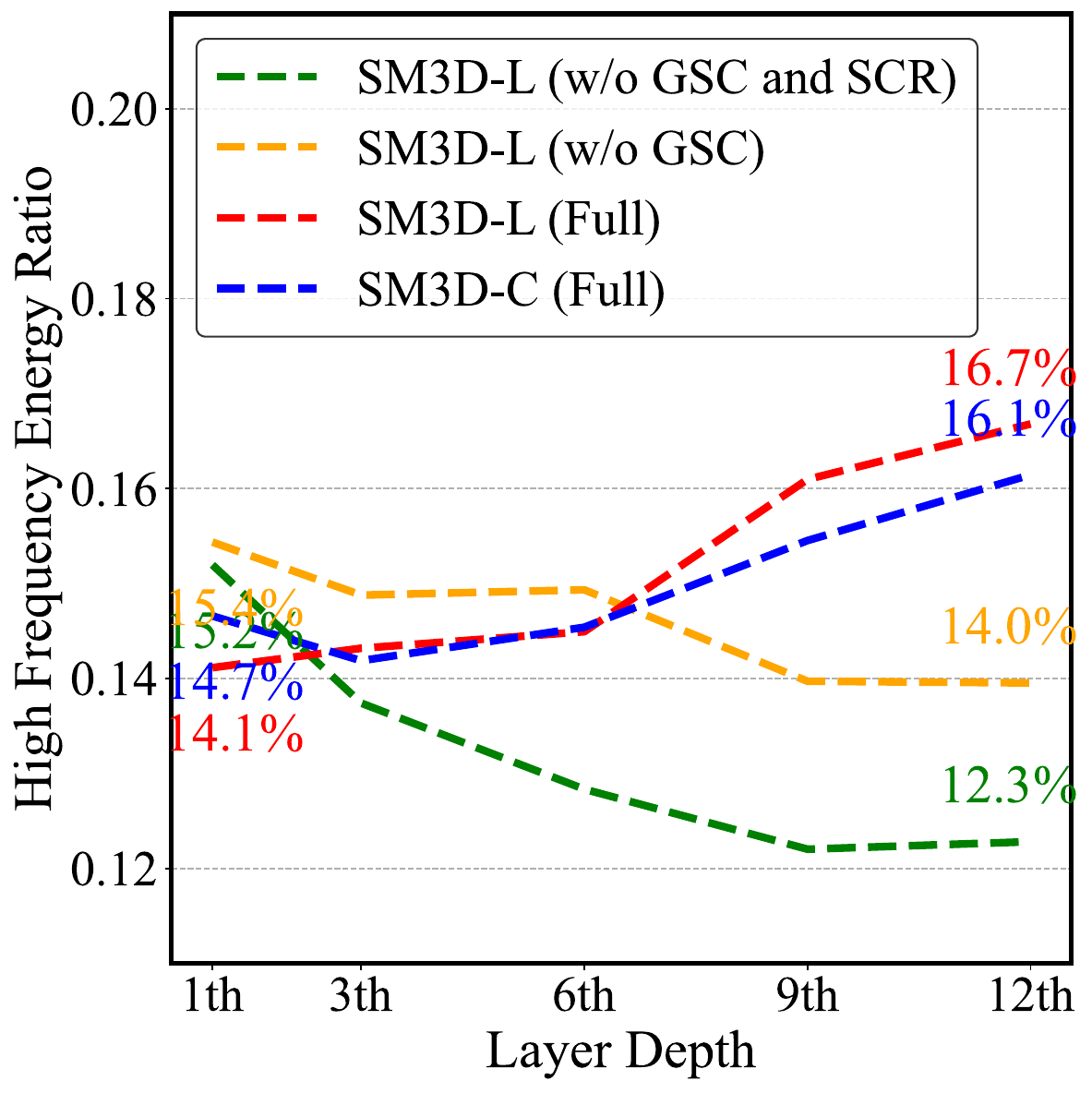}
		\label{fig ferq_high}	}
	\hfil 
	{\includegraphics[width=1.6in]{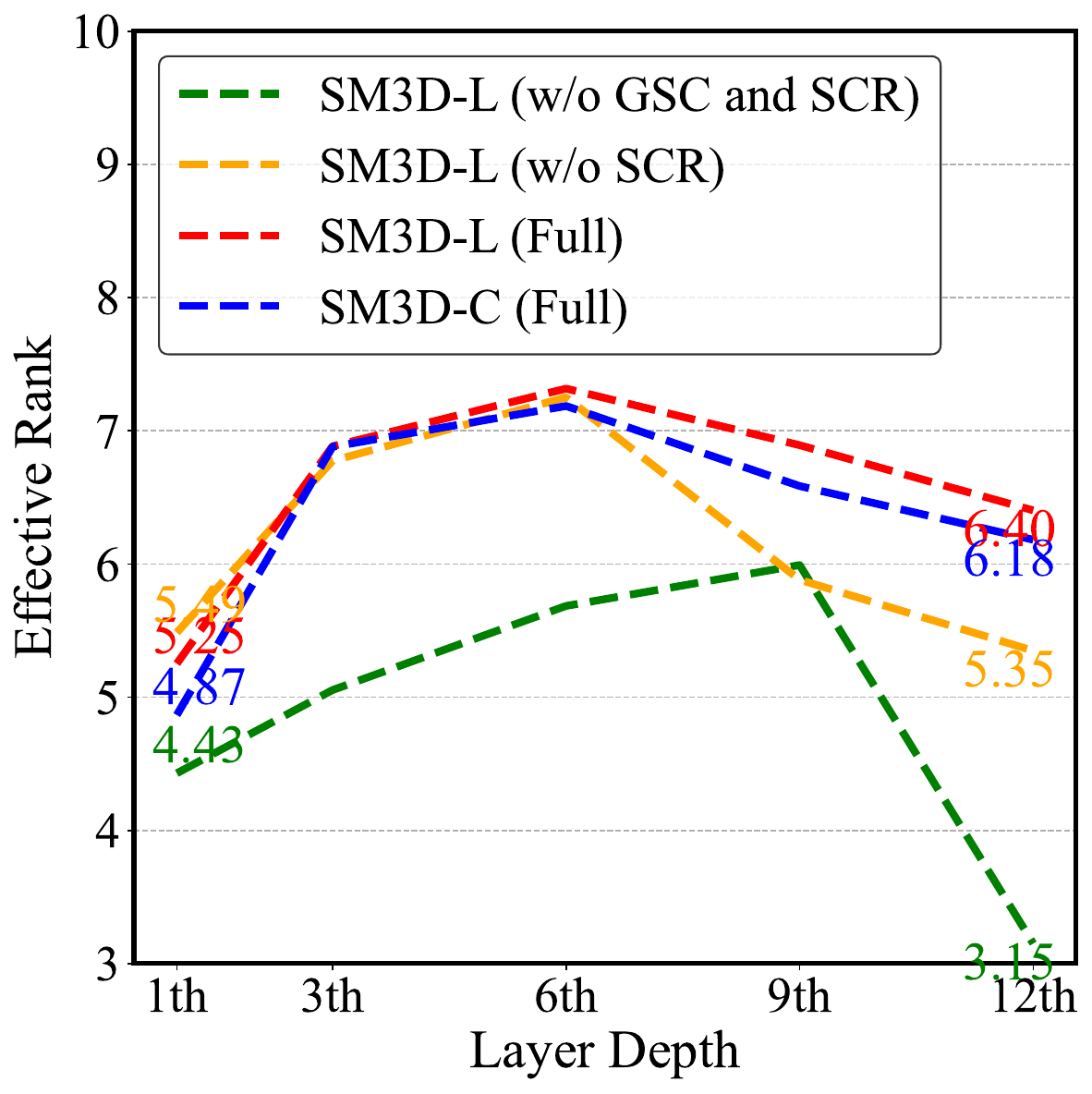}
		\label{fig rank}
	} 
	\caption{Layer-wise Spectral Evolution.} 
	\label{ferq}
\end{figure}

\textit{Spectral Evolution Analysis.}
As shown in Fig.\ref{ferq}, the Baseline's (w/o GSC and SCR) high-frequency energy decays from 0.152 to 0.122. This confirms the low-pass bias of standard SSMs. In contrast, SM3D-L and SM3D-C reverse this trend, showing a progressive increase in high-frequency energy. This behavior validates that GSC effectively preserves and reinforces the fine-grained geometric details.

Semantic dilution manifests as feature collapse. the Baseline's effective rank collapses drastically in the final layers, indicating severe semantic dilution. However, with the integration of SCR, SM3D-L and SM3D-Cmaintains a high effective rank. This demonstrates that the Global Spectral Anchor mechanism in SCR effectively rectifies feature drift, forcing the network to maintain semantically coherent feature manifold throughout deep propagation.

\vspace{-1mm}

\vspace{-2mm}
\section{Conclusion}
\label{Conclusion}

In this paper, we identify that the recursive nature of SSMs induces a spectral low-pass bias, which smooths critical geometric details and leads to semantic dilution during deep propagation. Accordingly, we propose SM3D, incorporating a Geometric Spectral Compensator (GSC) and a Semantic Coherence Refiner (SCR). GSC explicitly injects Laplacian-guided high-frequency residuals to restore structural fidelity, while SCR enforces global semantic consistency via spectral anchoring. By instantiating SCR with exact eigendecomposition (SM3D-L) and Chebyshev approximation (SM3D-C), SM3D bridges the gap between precision-critical tasks and large-scale scene understanding. Extensive experiments demonstrate the superiority of SM3D and verify its effectiveness in reversing the low-frequency feature bias inherent to standard SSMs. Future work will explore adaptive graph learning to optimize the spectral topology. We believe that a robust and interpretable spectral analysis can release Mamba's potential for efficient 3D perception.


\vspace{-1em}

\bibliographystyle{IEEEtran}
\bibliography{ref.bib}

@inproceedings{pointnet,
	author    = {Charles, R. Q. and Su, H. and Kaichun, M. and Guibas, L. J.},
	title     = {{PointNet}: Deep Learning on Point Sets for 3D Classification and Segmentation},
	booktitle = {Proc. IEEE Conf. on Comput. Vis. Pattern Recognit.},
	pages     = {77--85},
	year      = {2017}
}

@inproceedings{PointNet2,
	author    = {Qi, Charles R. and Yi, Li and Su, Hao and Guibas, Leonidas J.},
	title     = {{PointNet++}: Deep Hierarchical Feature Learning on Point Sets in a Metric Space},
	booktitle = {Adv. Neural Inf. Process. Syst.},
	pages     = {5105--5114},
	year      = {2017}
}

@inproceedings{SpiderCNN,
	author    = {Xu, Yifan and Fan, Tianqi and Xu, Mingye and Zeng, Long and Qiao, Yu},
	title     = {{SpiderCNN}: Deep Learning on Point Sets with Parameterized Convolutional Filters},
	booktitle = {Proc. Eur. Conf. Comput. Vis.},
	pages     = {90--105},
	year      = {2018}
}

@inproceedings{dgcnn,
	author    = {Wang, Yue and Sun, Yongbin and Liu, Ziwei and Sarma, Sanjay E. and Bronstein, Michael M. and Solomon, Justin M.},
	title     = {Dynamic Graph {CNN} for Learning on Point Clouds},
	booktitle = {{ACM} Trans. Graph.},
	volume    = {38},
	number    = {5},
	pages     = {146},
	year      = {2019}
}

@INPROCEEDINGS{s3dis,
	author={Armeni, Iro and Sener, Ozan and Zamir, Amir R. and Jiang, Helen and Brilakis, Ioannis and Fischer, Martin and Savarese, Silvio},
	booktitle = {Proc. IEEE Conf. on Comput. Vis. Pattern Recognit.},
	title={3D Semantic Parsing of Large-Scale Indoor Spaces}, 
	year={2016},
	pages={1534-1543},
}

@INPROCEEDINGS{grad-cam,
	author={Selvaraju, Ramprasaath R. and Cogswell, Michael and Das, Abhishek and others},
	booktitle = {Proc. IEEE Int. Conf. Comput. Vis.},
	title={Grad-CAM: Visual Explanations from Deep Networks via Gradient-Based Localization}, 
	year={2017},
	pages={618-626},
}

@article{swin3d, 
author = {Yu-Qi Yang and Yu-Xiao Guo and Jian-Yu Xiong and Yang Liu and Hao Pan and Peng-Shuai Wang and Xin Tong and Baining Guo},
title = {Swin3D: A pretrained transformer backbone for 3D indoor scene},
year = {2025},
journal = {Computational Visual Media},
volume = {11},
number = {1},
pages = {83-101},
}

@ARTICLE{Ortrga2018,
	author={Ortega, Antonio and Frossard, Pascal and Kovačević, Jelena and and others},
	journal={Proceedings of the IEEE}, 
	title={Graph Signal Processing: Overview, Challenges, and Applications}, 
	year={2018},
	volume={106},
	number={5},
	pages={808-828}}

@inproceedings{Pamba, 
	title={Pamba: Enhancing Global Interaction in Point Clouds via State Space Model},
	 volume={39},
	 number={5}, 
	booktitle = {Proc. AAAI Conf. Artif. Intell.},
	 author={Li, Zhuoyuan and Ai, Yubo and Lu, Jiahao and Wang, ChuXin and Deng, Jiacheng and Chang, Hanzhi and Liang, Yanzhe and Yang, Wenfei and Zhang, Shifeng and Zhang, Tianzhu}, 
	 year={2025}, 
	 month={Apr.}, 
	 pages={5092-5100} }

@inproceedings{KPConv,
	author    = {Thomas, Hugues and Qi, Charles R. and Deschaud, Jean-Emmanuel and Marcotegui, Beatriz and Goulette, Fran{\c{c}}ois and Guibas, Leonidas},
	title     = {{KPConv}: Flexible and Deformable Convolution for Point Clouds},
	booktitle = {Proc. IEEE Int. Conf. Comput. Vis.},
	pages     = {6410--6419},
	year      = {2019}
}

@inproceedings{modelnet,
	author    = {Wu, Zhirong and Song, Shuran and Khosla, Aditya and Yu, Fisher and Zhang, Linguang and Tang, Xiaoou and Xiao, Jianxiong},
	title     = {{3D ShapeNets}: A Deep Representation for Volumetric Shapes},
	booktitle = {Proc. IEEE Conf. on Comput. Vis. Pattern Recognit.},
	pages     = {1912--1920},
	year      = {2015}
}

@inproceedings{ScanObjectNN,
	author    = {Uy, Mikaela Angelina and Pham, Quang-Hieu and Hua, Binh-Son and Nguyen, Thanh and Yeung, Sai-Kit},
	title     = {Revisiting Point Cloud Classification: {A} New Benchmark Dataset and Classification Model on Real-World Data},
	booktitle = {Proc. IEEE Int. Conf. Comput. Vis.},
	pages     = {1588--1597},
	year      = {2019}
}

@inproceedings{DPC,
	author    = {Engelmann, Francis and Kontogianni, Theodora and Leibe, Bastian},
	title     = {Dilated Point Convolutions: {On} the Receptive Field Size of Point Convolutions on {3D} Point Clouds},
	booktitle = {Proc. IEEE Int. Conf. Robot. Autom.},
	pages     = {9463--9469},
	year      = {2020}
}

@inproceedings{Transformer,
	author    = {Vaswani, Ashish and others},
	title     = {Attention Is All You Need},
	booktitle = {Adv. Neural Inf. Process. Syst.},
	pages     = {6000--6010},
	year      = {2017}
}

@inproceedings{PatchFormer,
	author    = {Zhang, Cheng and Wan, Haocheng and Shen, Xinyi and Wu, Zizhao},
	title     = {{PatchFormer}: An Efficient Point Transformer with Patch Attention},
	booktitle = {Proc. IEEE Conf. on Comput. Vis. Pattern Recognit.},
	pages     = {11789--11798},
	year      = {2022}
}

@inproceedings{DSVT,
	author    = {Fan, Lue and Pang, Ziqi and Zhang, Tianyuan and Wang, Yu-Xiong and Zhao, Hang and Wang, Feng and Wang, Naiyan and Zhang, Zhaoxiang},
	title     = {Embracing Single Stride 3D Object Detector with Sparse Transformer},
	booktitle = {Proc. IEEE Conf. on Comput. Vis. Pattern Recognit.},
	pages     = {8448--8458},
	year      = {2022}
}

@inproceedings{ptv3,
	author    = {Wu, Xiaoyang and Jiang, Li and Wang, Peng-Shuai and Liu, Zhijian and Liu, Xihui and Qiao, Yu and Ouyang, Wanli and He, Tong and Zhao, Hengshuang},
	title     = {Point Transformer {V3}: Simpler, Faster, Stronger},
	booktitle = {Proc. IEEE Conf. on Comput. Vis. Pattern Recognit.},
	pages     = {4840--4851},
	year      = {2024}
}

@inproceedings{flatformer,
	author    = {Liu, Zhijian and Yang, Xinyu and Tang, Haotian and Yang, Shang and Han, Song},
	title     = {{FlatFormer}: Flattened Window Attention for Efficient Point Cloud Transformer},
	booktitle = {Proc. IEEE Conf. on Comput. Vis. Pattern Recognit.},
	pages     = {1200--1211},
	year      = {2023}
}

@inproceedings{spotr,
	author    = {Park, Jinyoung and Lee, Sanghyeok and Kim, Sihyeon and Xiong, Yunyang and Kim, Hyunwoo J.},
	title     = {Self-Positioning Point-Based Transformer for Point Cloud Understanding},
	booktitle = {Proc. IEEE Conf. on Comput. Vis. Pattern Recognit.},
	pages     = {21814--21823},
	year      = {2023}
}

@inproceedings{centerformer,
	author    = {Zhou and others},
	title     = {{CenterFormer}: Center-based Transformer for 3D Object Detection},
	booktitle = {Proc. Eur. Conf. Comput. Vis.},
	pages     = {496--513},
	year      = {2022}
}

@inproceedings{pointcont,
	author    = {Liu, Yahui and Tian, Bin and Lv, Yisheng and Li, Lingxi and Wang, Fei-Yue},
	title     = {Point Cloud Classification Using Content-Based Transformer via Clustering in Feature Space},
	booktitle = {{IEEE/CAA} J. Autom. Sinica},
	volume    = {11},
	number    = {1},
	pages     = {231},
	year      = {2024}
}

@inproceedings{Mamba,
	author    = {Gu, Albert and Dao, Tri},
	title     = {{Mamba}: Linear-Time Sequence Modeling with Selective State Spaces},
	booktitle = {First Conf. Lang. Modeling},
	year      = {2024}
}

@inproceedings{Gu2022,
	author    = {Gu, Albert and Goel, Karan and R{\'e}, Christopher},
	title     = {Efficiently Modeling Long Sequences with Structured State Spaces},
	booktitle = {Int. Conf. Learn. Represent.},
	year      = {2022}
}

@inproceedings{PointMamba,
	author    = {Liang, Dingkang and Zhou, Xin and Xu, Wei and Zhu, Xingkui and Zou, Zhikang and Ye, Xiaoqing and Tan, Xiao and Bai, Xiang},
	title     = {{PointMamba}: A Simple State Space Model for Point Cloud Analysis},
	booktitle = {Adv. Neural Inf. Process. Syst.},
	volume    = {37},
	pages     = {32653--32677},
	year      = {2024}
}

@inproceedings{PCM,
	author    = {Zhang, Tao and Yuan, Haobo and Qi, Lu and Zhang, Jiangning and Zhou, Qianyu and Ji, Shunping and Yan, Shuicheng and Li, Xiangtai},
	title     = {Point Cloud Mamba: Point Cloud Learning via State Space Model},
	booktitle = {Proc. AAAI Conf. Artif. Intell.},
	volume    = {39},
	number    = {10},
	pages     = {10121--10130},
	year      = {2025}
}

@inproceedings{Mamba3D,
	author    = {Han, Xu and Tang, Yuan and Wang, Zhaoxuan and Li, Xianzhi},
	title     = {{Mamba3D}: Enhancing Local Features for 3D Point Cloud Analysis via State Space Model},
	booktitle = {Proc. ACM Int. Conf. Multimed.},
	pages     = {4995--5004},
	year      = {2024}
}

@inproceedings{GridMamba,
	author    = {Yang, Yulong and Xun, Tianzhou and Hao, Kuangrong and Wei, Bing and Tang, Xuesong},
	title     = {{GridMamba}: Grid State Space Model for Large-Scale Point Cloud Analysis},
	booktitle = {Neurocomputing},
	volume    = {636},
	pages     = {129985},
	year      = {2025}
}

@inproceedings{pointramba,
	author    = {Wang, Zicheng and Chen, Zhenghao and Wu, Yiming and Zhao, Zhen and Zhou, Luping and Xu, Dong},
	title     = {{PoinTramba}: A Hybrid Transformer-Mamba Framework for Point Cloud Analysis},
	booktitle = {arXiv preprint arXiv:2405.15463},
	year      = {2024}
}

@article{DM3D,
	author  = {Liu, Bin and Wang, Chunyang and Liu, Xuelian},
	title   = {{DM3D}: Deformable Mamba via Offset-Guided Gaussian Sequencing for Point Cloud Understanding},
	journal = {arXiv preprint arXiv:2512.03424},
	year    = {2025}
}

@inproceedings{HydraMamba,
	author    = {Qu, Kanglin and Gao, Pan and Dai, Qun and Sun, Yuanhao},
	title     = {{HydraMamba}: Multi-Head State Space Model for Global Point Cloud Learning},
	booktitle = {Proc. ACM Int. Conf. Multimed.},
	pages     = {333--342},
	year      = {2025}
}

@inproceedings{Explor-lu2025,
	author    = {Lu, Dening and Gao, Kyle and Li, Jonathan and Zhang, Dedong and Xu, Linlin},
	title     = {Exploring Token Serialization for Mamba-Based LiDAR Point Cloud Segmentation},
	booktitle = {{IEEE} Trans. Geosci. Remote Sens.},
	volume    = {63},
	pages     = {1--14},
	year      = {2025}
}

@inproceedings{SAST,
	author    = {Bahri, Ali and Yazdanpanah, Moslem and Noori, Mehrdad and Dastani, Sahar and Cheraghalikhani, Milad and Hakim, Gustavo Adolfo Vargas and Osowiechi, David and Beizaee, Farzad and Ayed, Ismail Ben and Desrosiers, Christian},
	title     = {Spectral Informed Mamba for Robust Point Cloud Processing},
	booktitle = {Proc. IEEE Conf. on Comput. Vis. Pattern Recognit.},
	pages     = {11799--11809},
	year      = {2025}
}

@inproceedings{strumamba3d,
	author    = {Wang, Chuxin and Zha, Yixin and Yang, Wenfei and Zhang, Tianzhu},
	title     = {{StruMamba3D}: Exploring Structural Mamba for Self-supervised Point Cloud Representation Learning},
	journal   = {arXiv preprint arXiv:2506.21541},
	year      = {2025}
}

@inproceedings{point-mae,
	author    = {Pang, Yatian and Wang, Wenxiao and Tay, Francis E. H. and Liu, Wei and Tian, Yonghong and Yuan, Li},
	title     = {Masked Autoencoders for Point Cloud Self-supervised Learning},
	booktitle = {Proc. Eur. Conf. Comput. Vis.},
	pages     = {604--621},
	year      = {2022}
}

@inproceedings{Point-BERT,
	author    = {Yu, Xumin and Tang, Lulu and Rao, Yongming and others},
	title     = {{Point-BERT}: Pre-training 3D Point Cloud Transformers with Masked Point Modeling},
	booktitle = {Proc. IEEE Conf. on Comput. Vis. Pattern Recognit.},
	pages     = {19291--19300},
	year      = {2022}
}

@inproceedings{MaskPoint,
	author    = {Liu, Haotian and others},
	title     = {Masked Discrimination for Self-supervised Learning on Point Clouds},
	booktitle = {Proc. Eur. Conf. Comput. Vis.},
	pages     = {657--675},
	year      = {2022}
}

@inproceedings{Pointgpt,
	author    = {Chen, Guangyan and Wang, Meiling and Yang, Yi and Yu, Kai and Yuan, Li and Yue, Yufeng},
	title     = {{PointGPT}: Auto-regressively Generative Pre-training from Point Clouds},
	booktitle = {Adv. Neural Inf. Process. Syst.},
	volume    = {36},
	pages     = {29667--29679},
	year      = {2023}
}

@inproceedings{point-femae,
	author    = {Zha, Yaohua and Ji, Huizhen and Li, Jinmin and Li, Rongsheng and Dai, Tao and Chen, Bin and Wang, Zhi and Xia, Shu-Tao},
	title     = {Towards Compact 3D Representations via Point Feature Enhancement Masked Autoencoders},
	booktitle = {Proc. AAAI Conf. Artif. Intell.},
	volume    = {38},
	number    = {7},
	pages     = {6962--6970},
	year      = {2024}
}

@inproceedings{OcCo,
	author    = {Wang, H. and Liu, Q. and Yue, X. and Lasenby, J. and Kusner, M. J.},
	title     = {Unsupervised Point Cloud Pre-training via Occlusion Completion},
	booktitle = {Proc. IEEE Int. Conf. Comput. Vis.},
	pages     = {9762--9772},
	year      = {2021}
}

@inproceedings{ACT,
	author    = {Dong, Runpei and Qi, Zekun and Zhang, Linfeng and others},
	title     = {Autoencoders as Cross-Modal Teachers: Can Pretrained 2D Image Transformers Help 3D Representation Learning?},
	booktitle = {Int. Conf. Learn. Represent.},
	year      = {2023}
}

@inproceedings{point-pqae,
	author    = {Zhang, Xiangdong and Zhang, Shaofeng and Yan, Junchi},
	title     = {Towards More Diverse and Challenging Pre-training for Point Cloud Learning: Self-Supervised Cross Reconstruction with Decoupled Views},
	booktitle = {Proc. IEEE Int. Conf. Comput. Vis.},
	year      = {2025}
}

@inproceedings{PointNeXt,
	author    = {Qian, Guocheng and Li, Yuchen and Peng, Houwen and Mai, Jinjie and Hammoud, Hasan and Elhoseiny, Mohamed and Ghanem, Bernard},
	title     = {{PointNeXt}: Revisiting {PointNet++} with Improved Training and Scaling Strategies},
	booktitle = {Adv. Neural Inf. Process. Syst.},
	volume    = {35},
	pages     = {23192--23204},
	year      = {2022}
}

@inproceedings{APES,
	author    = {Wu, Chengzhi and Zheng, Junwei and Pfrommer, Julius and Beyerer, J{\"u}rgen},
	title     = {Attention-Based Point Cloud Edge Sampling},
	booktitle = {Proc. IEEE Conf. on Comput. Vis. Pattern Recognit.},
	pages     = {2814--2824},
	year      = {2023}
}

@article{ShapeNet,
	author  = {Chang, Angel and Funkhouser, Thomas and Guibas, Leonidas and others},
	title   = {{ShapeNet}: An Information-Rich 3D Model Repository},
	journal = {arXiv preprint arXiv:1512.03012},
	year    = {2015}
}

@ARTICLE{Laplacian,
	author={Shuman, David I and Narang, Sunil K. and Frossard, Pascal and others},
	journal={IEEE Signal Processing Magazine}, 
	title={The emerging field of signal processing on graphs: Extending high-dimensional data analysis to networks and other irregular domains}, 
	year={2013},
	volume={30},
	number={3},
	pages={83-98}}

@article{PointWavelet,
	author  = {Wen, Cheng and Long, Jianzhi and Yu, Baosheng and Tao, Dacheng},
	title   = {{PointWavelet}: Learning in Spectral Domain for 3-D Point Cloud Analysis},
	journal = {{IEEE} Trans. Neural Netw. Learn. Syst.},
	volume  = {36},
	number  = {3},
	pages   = {4400--4412},
	year    = {2025}
}

@inproceedings{Spectral-GANs,
	author    = {Ramasinghe, Sameera and Khan, Salman and Barnes, Nick and Gould, Stephen},
	title     = {{Spectral-GANs} for High-Resolution 3D Point-cloud Generation},
	booktitle = {Proc. {IEEE/RSJ} Int. Conf. Intell. Robots Syst.},
	pages     = {8169--8176},
	year      = {2020}
}

@article{pointgst,
	author  = {Liang, Dingkang and Feng, Tianrui and Zhou, Xin and others},
	title   = {Parameter-Efficient Fine-Tuning in Spectral Domain for Point Cloud Learning},
	journal = {{IEEE} Trans. Pattern Anal. Mach. Intell.},
	volume  = {47},
	number  = {12},
	pages   = {10949--10966},
	year    = {2025}
}

@article{S2ANet,
	author  = {Liu, Yujie and Sun, Xiaorui and Shao, Wenbin and Yuan, Yafu},
	title   = {{S2ANet}: Combining Local Spectral and Spatial Point Grouping for Point Cloud Processing},
	journal = {Virtual Reality \& Intell. Hardware},
	volume  = {6},
	pages   = {267--279},
	year    = {2024}
}

@article{DAM,
	author  = {Liu, Zhanwen and Cheng, Juanru and Fan, Jin and Lin, Shan and Wang, Yang and Zhao, Xiangmo},
	title   = {Multi-Modal Fusion Based on Depth Adaptive Mechanism for 3D Object Detection},
	journal = {{IEEE} Trans. Multimedia},
	volume  = {27},
	pages   = {707--717},
	year    = {2025}
}

@article{CliReg,
	author  = {Laserna, Javier and others},
	title   = {{CliReg}: Clique-Based Robust Point Cloud Registration},
	journal = {{IEEE} Trans. Robot.},
	volume  = {41},
	pages   = {1898--1917},
	year    = {2025}
}

@inproceedings{Weber2024,
	author    = {Weber, Maximilian and Wild, Daniel and Kleesiek, Jens and Egger, Jan and Gsaxner, Christina},
	title     = {Deep Learning-Based Point Cloud Registration for Augmented Reality-Guided Surgery},
	booktitle = {Proc. {IEEE} Int. Symp. Biomed. Imaging},
	pages     = {1--5},
	year      = {2024}
}

@inproceedings{XYScanNet,
	author    = {Liu, Hanzhou and Liu, Chengkai and Xu, Jiacong and Jiang, Peng and Lu, Mi},
	title     = {{XYScanNet}: A State Space Model for Single Image Deblurring},
	booktitle = {Proc. {IEEE/CVF} Conf. Comput. Vis. Pattern Recognit. Workshops},
	pages     = {770--780},
	year      = {2025}
}

@inproceedings{Wave-U-Mamba,
	author    = {Lee, Yongjoon and Kim, Chanwoo},
	title     = {{Wave-U-Mamba}: An End-To-End Framework For High-Quality And Efficient Speech Super Resolution},
	booktitle = {Proc. {IEEE} Int. Conf. Acoust. Speech Signal Process.},
	year      = {2025}
}

@inproceedings{lee2024,
	author    = {Lee, Seunghan and Hong, Juri and Lee, Kibok and Park, Taeyoung},
	title     = {Sequential Order-Robust Mamba for Time Series Forecasting},
	booktitle = {Adv. Neural Inf. Process. Syst.},
	year      = {2024}
}

\newpage
\appendix

\begin{figure*}[t]
	\centering
	\includegraphics[width=0.245\linewidth]{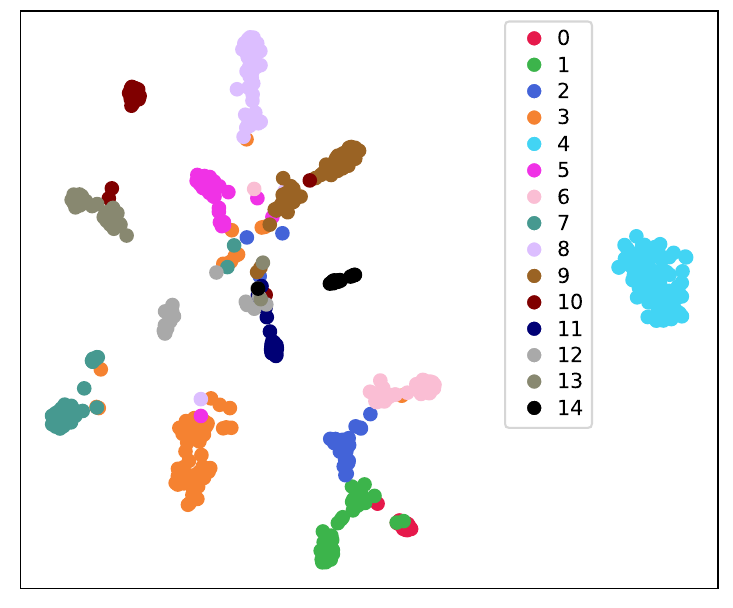}
	\hfill
	\includegraphics[width=0.245\linewidth]{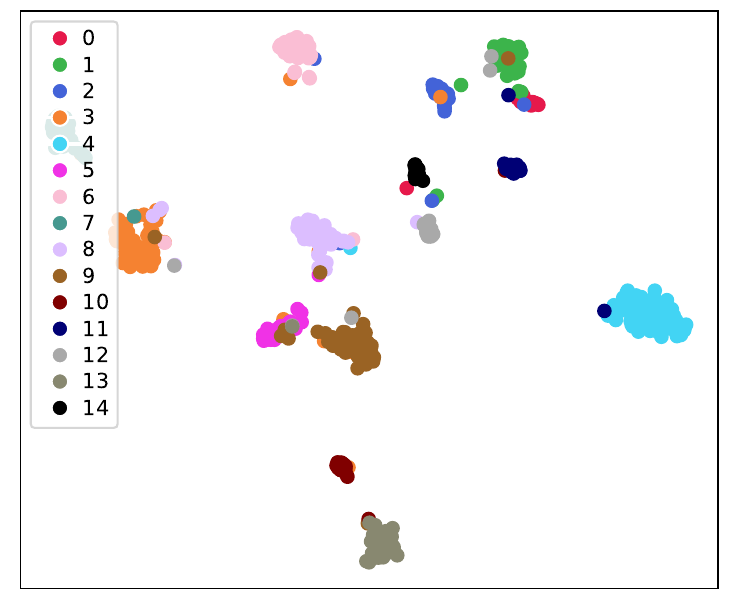}
	\hfill
	\includegraphics[width=0.245\linewidth]{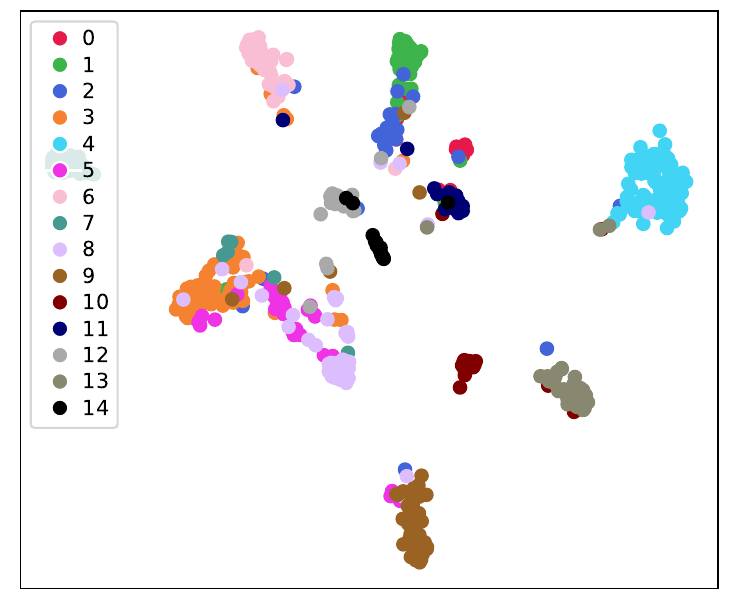}
	\hfill
	\includegraphics[width=0.245\linewidth]{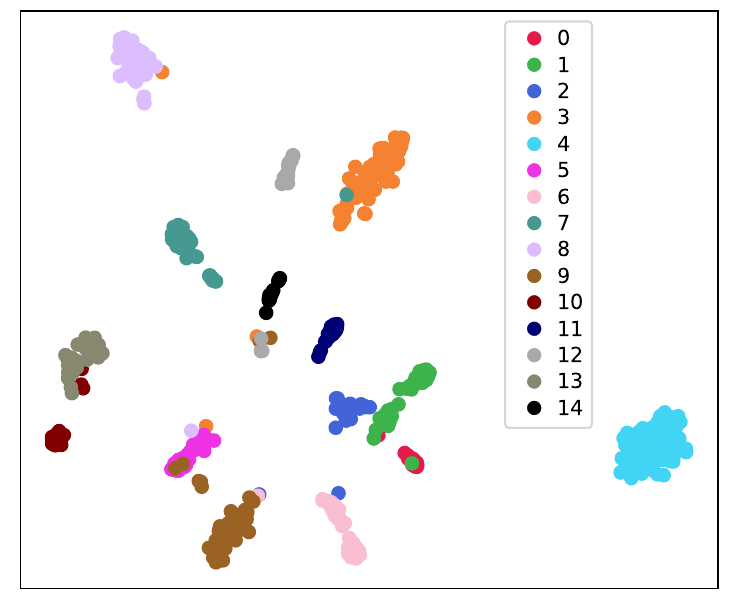}
	
	\vspace{-0.5em} 
	\subfloat[PointMamba(NeurIPS 24)]{\includegraphics[width=0.25\linewidth]{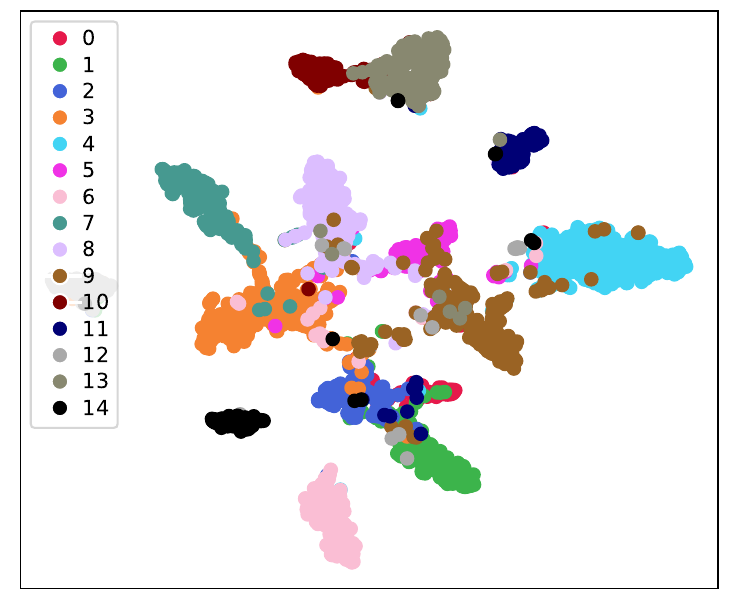}}
	\hfill
	\subfloat[Mamba3D (ACM MM 24)]{\includegraphics[width=0.25\linewidth]{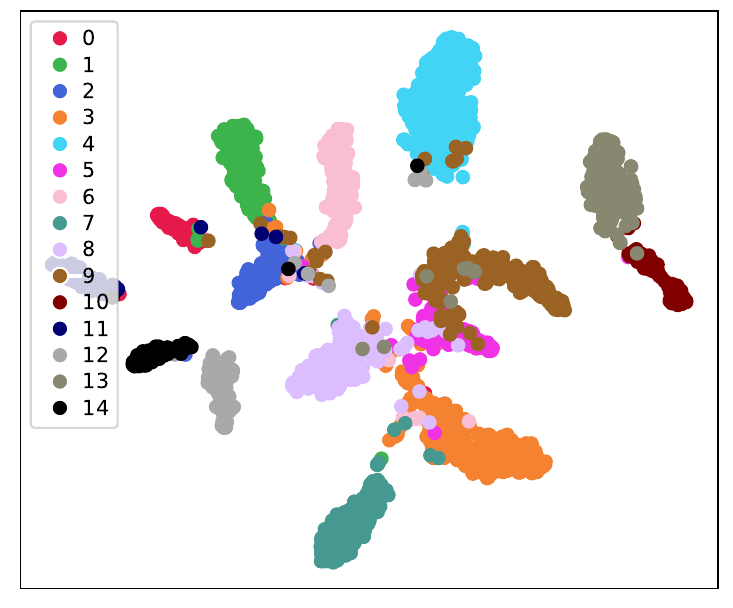}}
	\hfill
	\subfloat[Point-PQAE (ICCV 25)]{\includegraphics[width=0.25\linewidth]{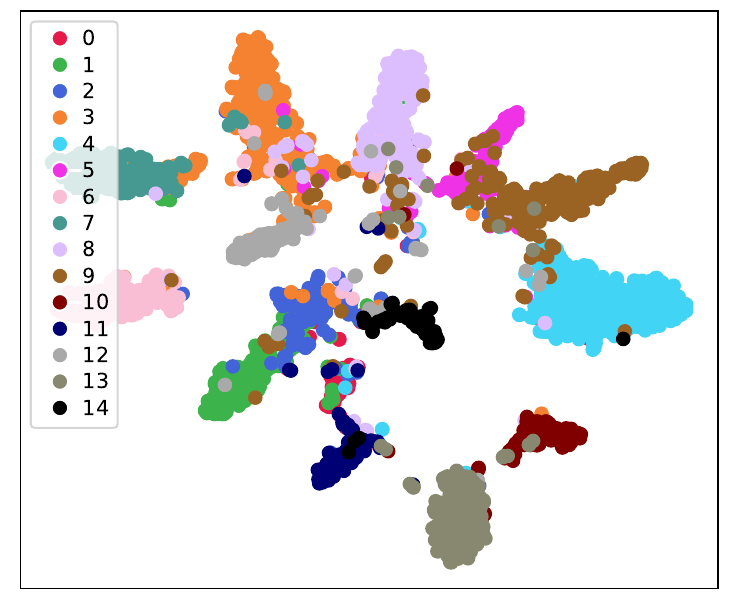}}
	\hfill
	\subfloat[SM3D-L (Ours)]{\includegraphics[width=0.25\linewidth]{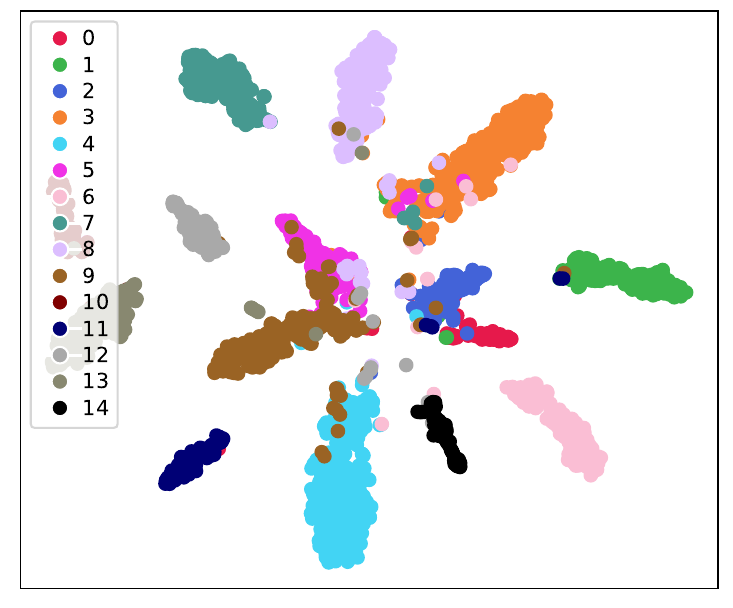}}
	
	\caption{t-SNE visualizations are shown for OBJ\_ONLY (top) and PB\_T50\_RS (bottom).}
	\label{tsne}
\end{figure*}

\subsection{More Results}
\textit{t-SNE visualization of Classification.} 
To qualitatively interpret these gains, we visualize the feature spaces via t-SNE. As shown in Fig.\ref{tsne}, while PointMamba and Mamba3D exhibit scattered distributions with noticeable inter-class overlap, SpecMamba3D produces highly compact and well-separated clusters. This visual evidence corroborates that rectifying spectral bias and semantic dilution leads to more robust and discriminative representations in the deep space.

\textit{Analysis of parameters.} The model peak performance at a neighborhood of $K_G=4$. Smaller values ($K_S=3$) lack sufficient geometric context, while larger values ($K_G \ge 6$) introduce redundant spatial smoothing, diluting the sharp high-frequency residuals we aim to capture. For the SCR-C module, $K_S$ controls the density of the global graph used for Chebyshev approximation. Performance improves as $K_S$ increases from 8 to 32, as a denser graph provides better global connectivity for spectral propagation. However, increasing $K_S$ further to 64 yields diminishing returns and higher computational cost, making $K_S=32$ the optimal trade-off for effective global anchoring.

\begin{table}[h]
	\centering
	\caption{\textbf{Ablation on key parameters.} $K_G$: Local neighborhood size in GSC. $K_S$: Graph neighbor count in SCR-C.}
	
	\begin{minipage}[t]{0.48\linewidth}
		\centering
		(a) Effect of GSC $K_G$
		\small
		\begin{tabular}{cc}
			\toprule
			$K_G$ Value & PB\_T50\_RS    \\ \midrule
			3           & 91.04          \\
			\textbf{4}  & \textbf{91.11} \\
			6           & 91.09          \\
			8           & 91.07          \\ \bottomrule
		\end{tabular}
	\end{minipage}
	\hfill
	\begin{minipage}[t]{0.48\linewidth}
		\centering
		(b) Effect of SCR $K_S$
		\small
		\begin{tabular}{cc}
			\toprule
			$K_S$ Value & PB\_T50\_RS    \\ \midrule
			8           & 91.03          \\
			16          & 91.71          \\
			\textbf{32} & \textbf{92.37} \\
			64          & 92.29          \\ \bottomrule
		\end{tabular}
	\end{minipage}
	
	\label{tab:param}
\end{table}

\begin{figure}[!t] 
	\centering
	\subfloat[Random and geometric high frequency masking strategies.]{\includegraphics[width=1.6in]{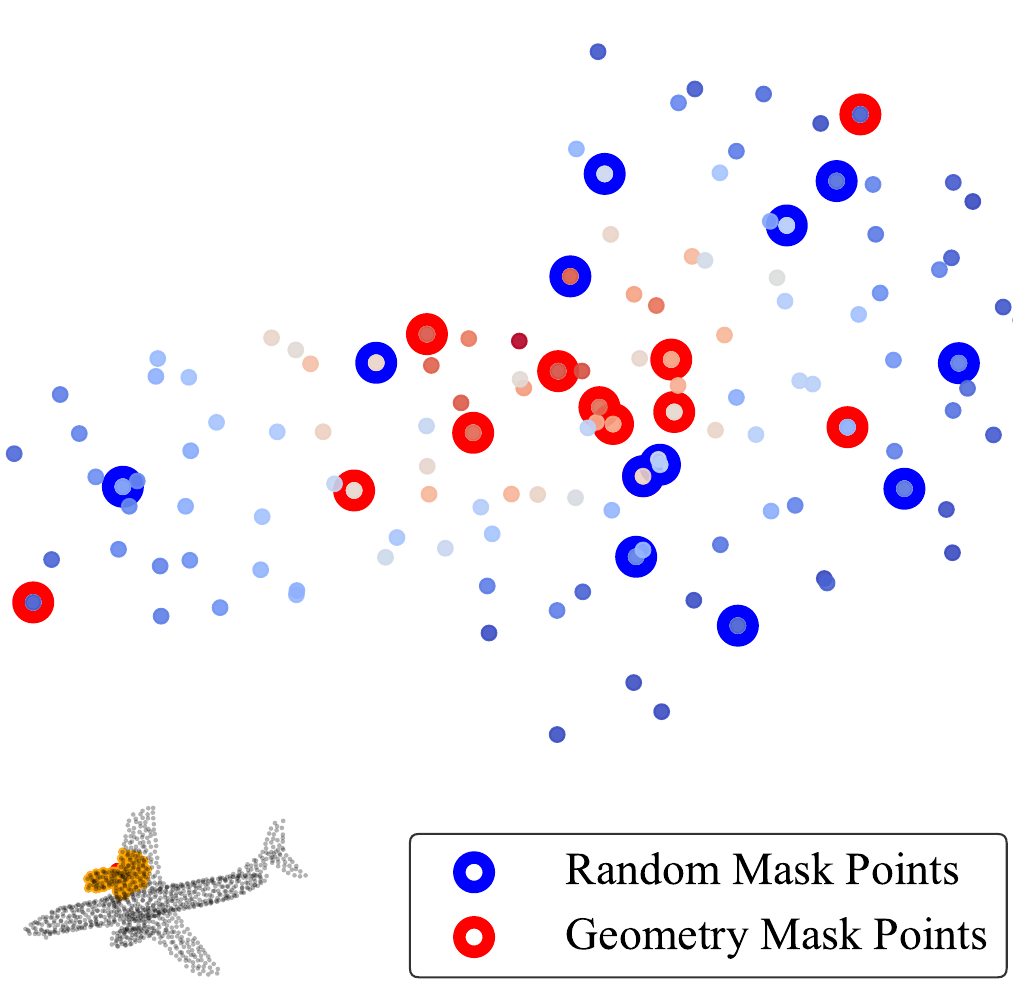}}
	\label{mask1}
	\hfil
	\subfloat[OA under different masking ratios.]{\includegraphics[width=1.6in]{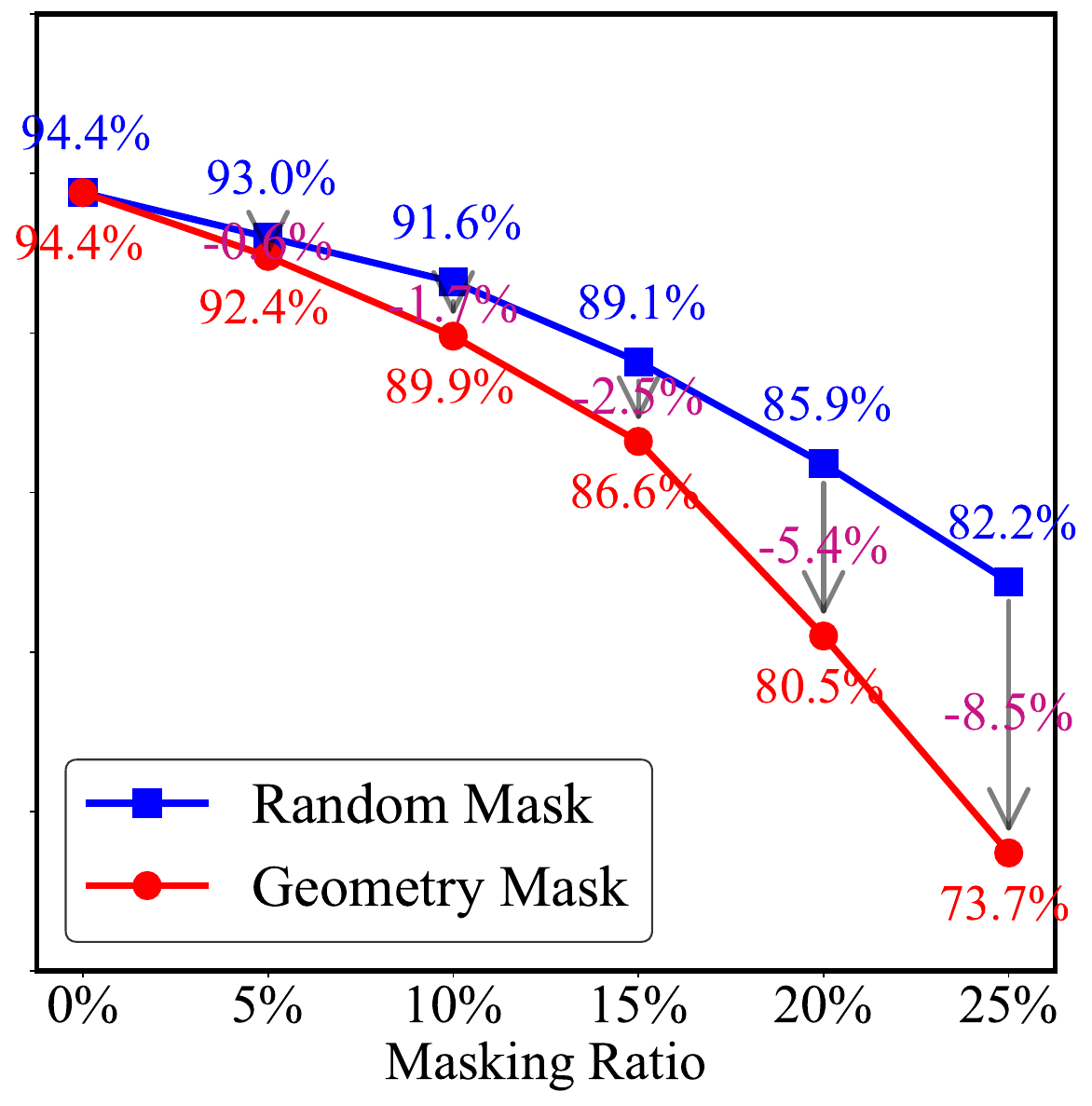}}
	\label{mask2}
	
	\subfloat[Accuracy degradation before and after masking.]{\includegraphics[width=1.6in]{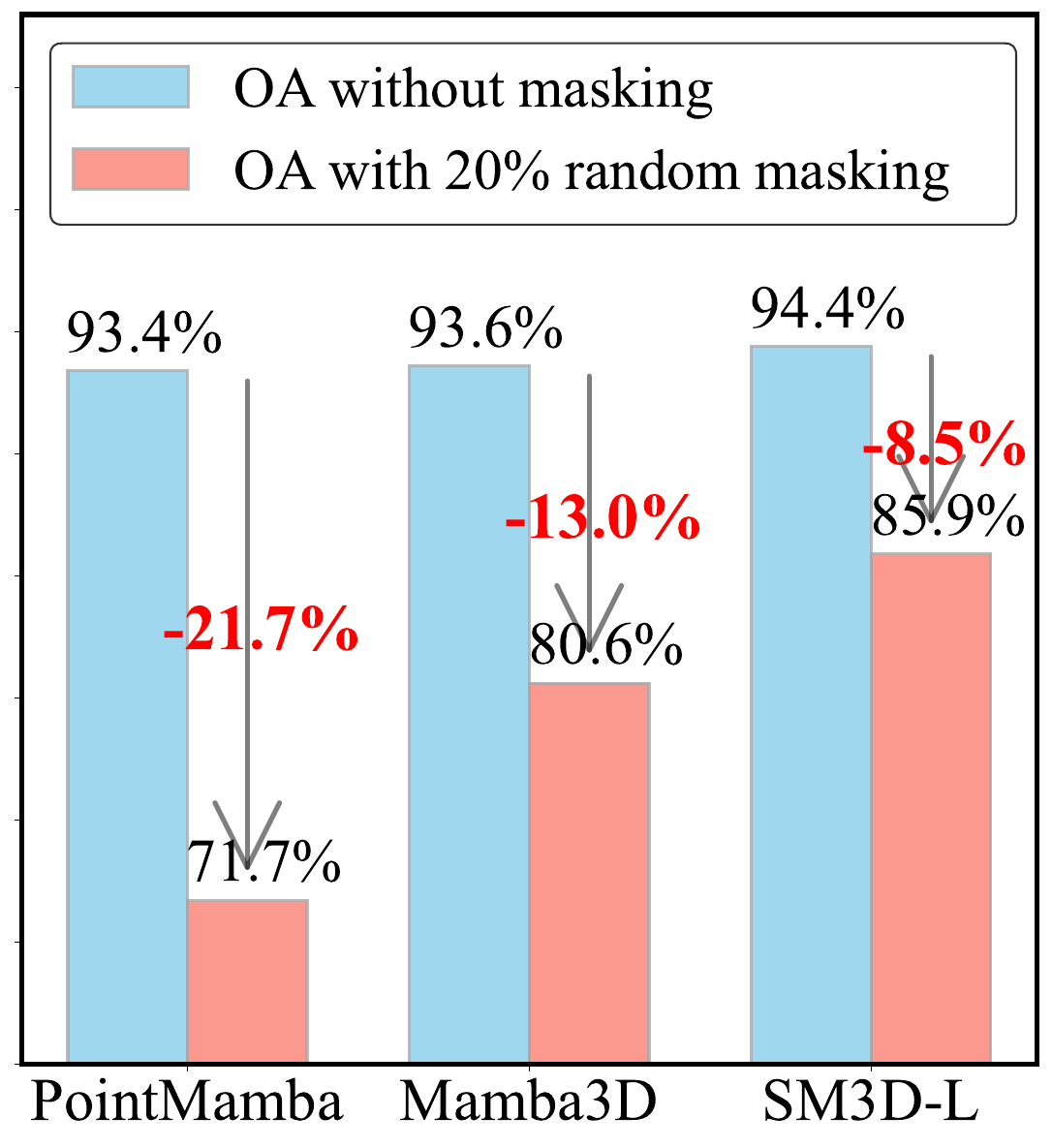}}
	\label{mask3}
	\hfil
	\subfloat[Effective rank (ER) variation before and after masking.]{\includegraphics[width=1.6in]{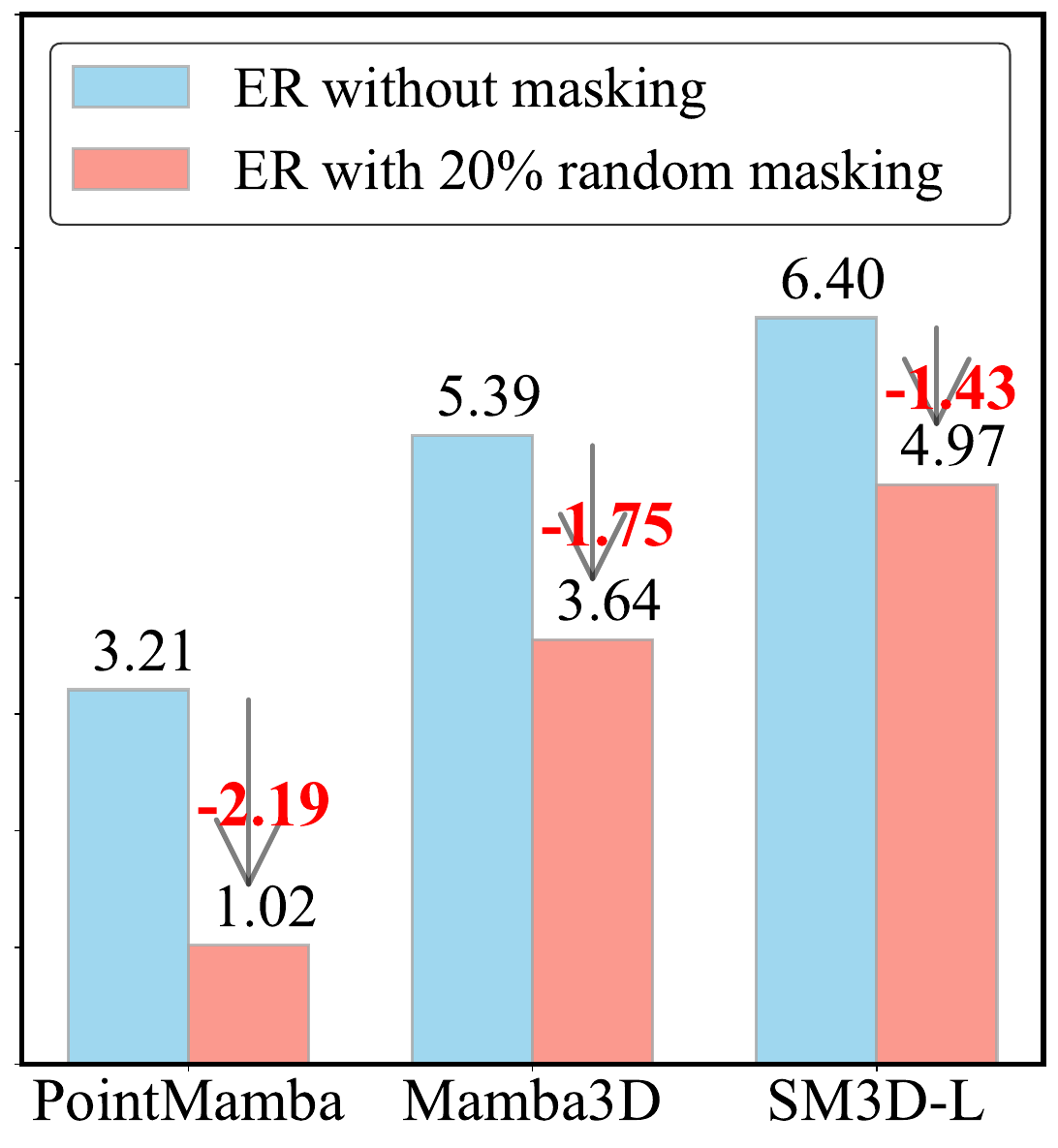}}
	\label{mask4}
	\caption{Layer-wise Spectral Evolution.} 
	\label{mask}
\end{figure}

\subsection{Masking Experiment}

To causally evaluate the role of high-frequency geometric information, we conduct targeted masking experiments, as illustrated in Fig.\ref{mask}. 

To causally validate the role of high-frequency geometric information, we conduct targeted masking experiments on ModelNet40, as illustrated in Fig.\ref{mask1}. We compare two strategies:random masking, which uniformly removes points irrespective of geometry, and geometric high-frequency masking, which preferentially removes points with strong Laplacian high-frequency responses. Both strategies remove an identical proportion of points to ensure a fair comparison.

\textit{Impact on Performance.}
As shown in Fig.\ref{mask2}, geometric high-frequency masking induces a significantly steeper performance decline. At a 25\% masking ratio, the accuracy drop under geometric high-frequency masking is 8.5\% greater than under random masking. This pronounced gap demonstrates that high-frequency geometric regions encode the most discriminative structural cues and cannot be compensated by low-frequency global context alone.

Under 20\% random masking, SM3D only drops by 8.2\%, lower than PointMamba and Mamba3D. This robustness stems from the GSC module, which actively reinforces stable high-frequency spectral features. Even when partial geometry is corrupted, these injected spectral residuals provide sufficient evidence to sustain correct predictions.

\textit{Feature Stability.} To quantify the impact of masking on semantic, we analyze the ER of the final layer's feature tokens. As shown in Fig.\ref{mask3}, the ER of PointMamba collapses severely from 3.21 to 1.02, indicating a drastic dilution of semantic information. In contrast, SM3D-L maintains an ER of 4.97 even after masking. This stability is primarily attributed to the SCR module’s global spectral anchoring mechanism, which ensures that the latent representation remains semantically coherent and resists feature collapse.

\vfill

\end{document}